\pdfoutput=1

\documentclass[11pt]{article}

\usepackage{ACL2023}

\usepackage{times}
\usepackage{latexsym}

\usepackage[T1]{fontenc}

\usepackage[utf8]{inputenc}

\usepackage{microtype}

\usepackage{inconsolata}

\usepackage{url}            \usepackage{booktabs}       \usepackage{amsfonts}       \usepackage{nicefrac}       \usepackage{microtype}      \usepackage{xcolor}         \usepackage{array}
\usepackage{enumitem}

\usepackage{microtype}
\usepackage{graphicx}
\usepackage{subcaption}
\usepackage{booktabs} \usepackage{floatrow}
\usepackage{algorithm2e}
\usepackage{tikz}
\usetikzlibrary{calc}

\usepackage{multirow}

\usepackage{amsmath}
\usepackage{amssymb}
\usepackage{mathtools}
\usepackage{amsthm}
\usepackage{arydshln}

\title{Hidden Schema Networks}

\author{
Rams\'es J. S\'anchez\textsuperscript{1,2*}, Lukas Conrads\textsuperscript{1,2}, Pascal Welke\textsuperscript{1,2,3}, \\ \textbf{Kostadin Cvejoski\textsuperscript{1,4}}, \textbf{C\'esar Ojeda\textsuperscript{5}} \\
\textsuperscript{1}Lamarr Institute, 
\textsuperscript{2}University of Bonn, 
\textsuperscript{3}TU Wien,
\textsuperscript{4}Fraunhofer IAIS \\
\textsuperscript{5} University of Potsdam
}

\def\*#1{\mathbf{#1}}
\def\+#1{\boldsymbol{#1}}

\begin{document}
\maketitle
\begin{abstract}
Large, pretrained language models infer powerful representations that encode rich semantic and syntactic content, albeit implicitly. In this work we introduce a novel neural language model that enforces, via inductive biases, explicit relational structures which allow for compositionality onto the output representations of pretrained language models.
Specifically, the model encodes sentences into sequences of symbols (composed representations), which correspond to the nodes visited by biased random walkers on a global latent graph, and infers the posterior distribution of the latter.
We first demonstrate that the model is able to uncover ground-truth graphs from artificially generated datasets of random token sequences. Next, we leverage pretrained BERT and GPT-2 language models as encoder and decoder, respectively, to infer networks of symbols (schemata) from natural language datasets. Our experiments show that (i) the inferred symbols can be interpreted as encoding different aspects of language, as e.g. topics or sentiments, and that (ii) GPT-like models can effectively be conditioned on symbolic representations.
Finally, we explore training autoregressive, random walk ``reasoning" models on schema networks inferred from commonsense knowledge databases, and using the sampled paths to enhance the performance of pretrained language models on commonsense \textit{If-Then} reasoning tasks.
\end{abstract}

\section{Introduction}
Much of the developmental and causal theories of human cognition are predicated on \textit{relational} structures of knowledge that naturally
exhibit 
\textit{compositionality}. 
Semantic content is intrinsically relational, as one is only able to explain a given unit of knowledge -- such as a concept, word or perception -- insofar as there are other units of knowledge which relate to it \citep{block1986advertisement, margolis1999concepts}.
Thus we can partially construe a concept through its relationships to other concepts (like when we say ``a dog is an animal that barks''),
just as 
we can partially construe it through its relationships to our perceptions (when we say ``\textit{that} is a dog'', whilst pointing to a dog on the street) 
or the words we use (when we use the word \textit{dog} to refer to the concept \textit{dog}).
Likewise,
we can partially construe words
not only through their relationships to concepts or percepts, but also 
through their relationships to other words,
as words that occur in the same context tend to have similar meanings \citep{harris1954distributional, firth1957synopsis}.
Note that is precisely this contextual semantic word content what we explicit have access to, when processing our raw text datasets.
On the other hand, 
generalization, reasoning and understanding seem to be inevitably tied to the compositional nature of knowledge. 
Indeed, the ability to compose a set of knowledge units (and their relations) into new, more complex relational units, which can be deployed to understand and reason about unseen data
-- a feature usually referred to as combinatorial generalization -- is regarded as key to human-level intelligence \citep{fodor1988connectionism, fodor2002compositionality, lake2017building, battaglia2018relational}.
Relational structures allowing for compositionality thus seem to
comprise not a sufficient, but a necessary attribute of any representation scheme that strives for the generalization power of human cognition.

From the computational side, if one is to inform any modern language model 
with such structural characteristics, one will initially encounter the problem of finding suitable primitives or data structures. 
Distributed continuous word representations \citep{bengio2003neural}, for example, are routinely leveraged in many different downstream tasks.
These representations are trained to encode average contextual semantics 
-- precisely the kind of semantic content typical of word co-occurrence relations we mentioned above --
into a semantic space, which allows meaning to change continuously within it \citep{mikolov2013efficient}.
Yet, despite earlier attempts \citep{mitchell2008vector}, it is unclear whether such representations can be meaningfully composed into representations of, say, unseen sentences and thus mimic the compositional character of natural language.
More recently, contextualized continuous word representations inferred by deep learning architectures have shown spectacular results in many NLP tasks \citep{radford2018improving, devlin2018bert, radford2019language, brown2020language}. 
Their success stems from those models' ability to infer flexible representations through, \textit{inter alia}, raw, massive datasets, data-scalable attention mechanisms and minimal inductive biases \citep{vaswani2017attention}.
These representations are known to not only contain rich contextual word semantics, but also consistently encode sentence-level grammar \citep{hewitt2019structural}, and 
the models from which they are obtained have been shown to implement different notions of compositionality too \citep{hupkes2020compositionality, wei2022chain}. 
Nevertheless, it is still unclear whether such representations can be composed into representations of novel sentences \citep{yu-ettinger-2020-assessing, bhathena-etal-2020-evaluating}. 
In fact, most of their syntactic properties are implicit and therefore inferred only a posteriori, typically through probes which neither guarantee their presence, nor establish how they were obtained in the first place \citep{liu-etal-2019-linguistic, rogers-etal-2020-primer, ravichander-etal-2021-probing}.

Roughly following the program outlined by \citet{tenenbaum2011grow}, 
we develop a novel, neural language model -- the Hidden Schema Network model (HSN) -- that enforces, via inductive biases, a discrete, \textit{relational} structure for sentence representation 
which allows for \emph{compositionality}\footnote{Note that throughout the paper we refer only to \textit{compositionality of representations} and not to the compositional functions that can be implemented by the models we use. The latter, functional compositionality, is studied by e.g.~\citet{hupkes2020compositionality}.} onto the output representations
of large, pretrained language models (LPLM).
Using a variational autoencoder (VAE) framework \citep{kingma2013auto, rezende2014stochastic}, HSN leverages LPLMs to encode natural language sentences into sequences of symbols, which correspond to the nodes visited by biased random walkers on a global latent graph, while inferring the posterior distribution of the latter.

In practice, 
translating the implicit knowledge encoded by LPLMs into explicit relational structures
has some naturally appealing properties. 
For example, HSN can support symbolic reasoning via inference of missing semantic connections, through high-order paths along the inferred graphs \citep{lao-etal-2011-random, das2018go, chen2020review}. 
Likewise secondary, autoregressive ``reasoning'' models can be trained on the symbol sequences inferred by HSN from task-specific natural language sequences, like e.g. 
question-answer pairs.
The reasoning paths sampled from such autoregressive models could then be used to inform GPT-like models and improve their performance on natural language understanding tasks, with which they are known to struggle \citep{brown2020language, dunietz-etal-2020-test, talmor2021commonsenseqa}.
Below we demonstrate that 
(i) HSN is able to uncover ground-truth graphs from artificially generated datasets of random token sequences, and that
(ii) using pretrained BERT and GPT-2 language models as encoder and decoder, respectively, HSN is able to infer schema networks from natural language datasets, whose symbols encode different aspects of language (like e.g.~topics or sentiments).
Finally, we also explore training secondary, autoregressive models on the schema networks inferred from commonsense knowledge databases, and using the sampled paths to enhance the performance of LPLM on commonsense \textit{If-Then} reasoning tasks.

 \label{sec:introduction}

\section{Related Work}
\begin{figure*}[t!]
    \centering
    \includegraphics[width=.9\textwidth]{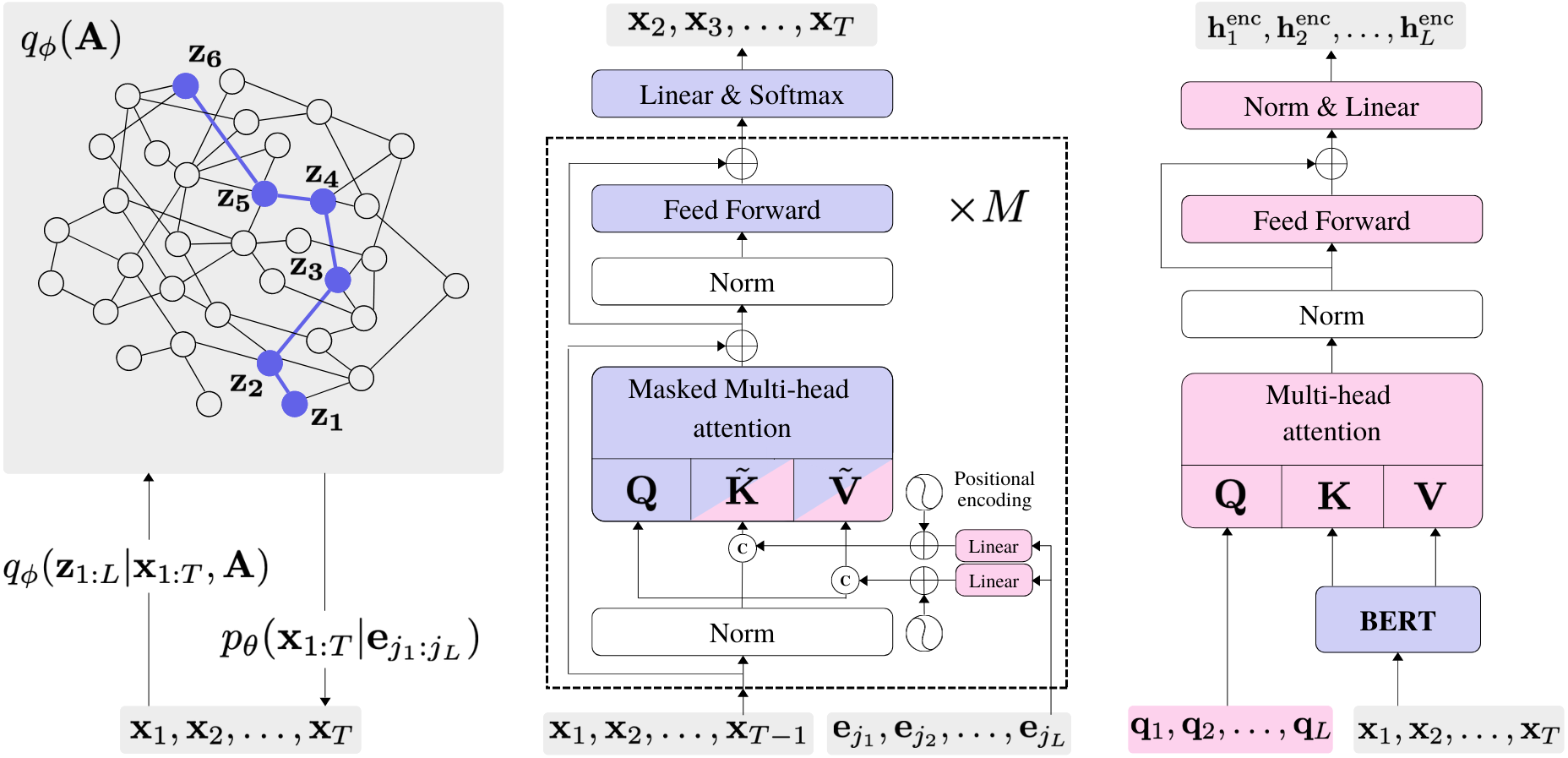}
    \caption{\textbf{Left:} Hidden Schema Network model. \textbf{Center:} Decoder architecture as a modified GPT-2 with $M$ layers and pseudo-self-attention mechanism to attend to the schema $\*e_{j_1:j_L}$. This schema is defined as a sequence of $L$ integers, each of which is represented as a one-hot vector.
The ``c'' operations labels concatenation. \textbf{Right}: Encoder architecture as BERT, followed by a single Transformer block. In both center and right figure purple (pink) shaded blocks represent submodules with pretrained (randomly initialized) parameters. 
    }
    \label{fig:model}
\end{figure*}

In cognitive psychology, a schema is (roughly) defined as a large, complex unit of knowledge representing what is typical of a group of instances \citep{bartlett-remembering, piaget1948langage, rumelhart2017schemata}. 
Marvin Minsky's frames \citep{minsky1974framework, minsky1975minsky} are similar in function to a schema, but perhaps more easily characterized in terms of data structures.
We use these terms in a loose fashion, however. 
Our aim being only
to be suggestive of the general problem of knowledge representation \citep{thagard1984frames}.
We are in fact concerned with representation schemes for natural language processing. Within the context of linguistics, \citet{jackendoff1978argument} argues that there must be a level of representation -- the so-called conceptual structures -- at which information conveyed by language must be compatible with information coming from sensory systems. 
Conceptual structures must, he goes on, be able to represent all the conceptual distinctions made by natural language, and provide some degree of compositionality. 
Earlier computational models implementing (some kind of) conceptual structure rely on either hand-coded (semantic) network representations \citep{quillan1966semantic, collins1969retrieval, brachman1977s} or hand-coded databases \citep{mcclelland2003parallel}.
Other works focus instead on learning semantic representations from raw text data directly, via topic models \citep{griffiths2007topics},
and even infer latent concept graphs through nonparametric priors \citep{NIPS2010_6c8dba7d}.
In sharp contrast with these works, (a line of) modern, neuro-symbolic reasoning approaches in NLP seek to combine the implicit knowledge of LPLM with external information, structured as knowledge graphs, in order to solve commonsense question-answering tasks \citep{lin-etal-2019-kagnet, ma-etal-2019-towards, yasunaga-etal-2021-qa, verga-etal-2021-adaptable, kapanipathi-etal-2021-leveraging, hu2022empowering}.
Note that the knowledge graphs here can also be interpreted as (some class of) human-authored conceptual structure (see however \citet{west2021symbolic}, who shows that LPLM can generate commonsense knowledge graphs too).
Our work runs perpendicular to these approaches, for it leverages LPLMs to automatically discover, in an unsupervised fashion, knowledge graphs in representation space, and introduces reasoning via random walk processes conditioned on the discovered graphs.
This problem necessarily involves the (neural) inference of discrete variables, which has also been successfully treated in the past \citep{NIPS2017_7a98af17, 10.5555/3305381.3305542, zhao-etal-2018-unsupervised, kaiser2018discrete, pmlr-v80-kaiser18a, pmlr-v119-stratos20a}.

 \label{sec:related_work} 

\section{Hidden Schema Networks}

We address the problem of learning the joint probability distribution over sequences of words, while inferring symbolic representations capturing their semantics.  
Neural autoregressive language models approximate such distributions with a product over conditional probabilities, such that
\begin{equation}
    p(\*x_{1:T}) = \prod_{i=1}^T p_{\theta}(\*x_i | \*x_{<i}),
    \label{eq:language_model}
\end{equation}
where $\*x_{1:T} = (\*x_1, \*x_2, \dots, \*x_T)$ labels the sequence of words in question, 
and each conditional is given by (the pdf of) a categorical distribution over some vocabulary of size $V$.
The class probabilities of these
conditionals
are generally computed as 
$
    \+\pi_i = \text{softmax}(  \* W \cdot \*h_{\theta}(\*x_{<i}) ),
$
with $\*W \in \mathbb{R}^{V \times D}$ trainable, and $D$ the output dimension of $\*h_{\theta}$, a deep neural network model with parameter set $\theta$ \citep{bengio2003neural}.
Models of this form allow for tractable estimation of and sampling from either the joint distribution, or any product of the conditionals in Eq.~\ref{eq:language_model}.
Indeed, their recent implementation in terms of large-capacity, self-attention architectures, the likes of GPT-3 \citep{brown2020language}, has been shown to generate syntactically correct, diverse and fluent text.
In what follows we attempt to condition the joint distribution of Eq.~\ref{eq:language_model} on an additional
latent, discrete representation which can, at least in principle, capture the relational and compositional features of semantic content.

Let us assume there is a set of $K$ symbols, ${\cal E} = \{\*e_1, \*e_2, \dots, \*e_K\}$, each of which encode some high-level, abstract semantic content of natural language. 
Let this set be the set of nodes of a hidden (semantic) graph $\cal G$, with adjacency matrix $\*A$, so that adjacent (connected) symbols are semantically related.
These symbols can generically be defined as learnable, dense vectors in $\mathbb{R}^S$, for some dimension $S$.
Without loss of generality, however, we opt below for simple indicator (``one-hot'') vectors of dimension $K$ instead.
We define a \textit{schema} $\*e_{j_1:j_L}$ as a sequence of $L \ll K$ symbols $(\*e_{j_1}, \*e_{j_2}, \dots, \*e_{j_L})$, where the indices $j_1, \dots, j_L$ label a subset of connected nodes in $\cal G$.
Accordingly, we refer to $\cal G$ as a \textit{schema network}.
The symbols composing the schemata are chosen through a $L$-step stochastic process conditioned on $\cal G$.
Partially motivated by research on random walks and human memory search \citep{griffiths2007google, abbott2012human}, 
as well as by the simplicity of their inference, 
we choose to compose the schemata via biased random walk processes on $\cal G$, 
and leave exploring different schema processes for future work.
Finally, let us assume that natural language sentences are generated conditioned on these schema representations.
We can then define a generative language model as follows.

\subsection{Generative Model}
\label{sec:gen_model}

Let the joint probability over a sequence $\*x_{1:T}$ of $T$ words, together with the hidden graph $\cal G$, be
\begin{multline}
    p_{\theta}(\*x_{1:T}, \*A) = \\ \sum_{\*z_{1:L}} p_{\theta}(\*x_{1:T} | \*e_{j_1:j_L}) p(\*z_{1:L} | \*A) p(\*A),
    \label{eq:generative_model}
\end{multline}
where $\*z_{1:L}$ labels the sequence $\*z_1, \dots, \*z_L$ of $K$-dimensional, one-hot vectors representing the node labels $j_1, \dots, j_L$ visited by a random walker on $\cal G$, and $\theta$ denotes the trainable model parameters.
Note that we introduced the one-hot representation of $j_i$ for notational convenience, as shall become evident below\footnote{Explicitly, $j_i$ denotes the index of the non-zero component of $\*z_i$, i.e. $j_i = \{k \in [1, K]: z_i^k = 1\}$, with the superindex $k$ denoting the components of $\*z_i$.}.
Next, we specify the different components of Eq.~\ref{eq:generative_model}.

 \textbf{Prior over (global) graph}. A prior on the adjacency matrix $p(\*A)$ allows us to control the topological properties of $\cal G$. 
One can choose, for example, random graph models whose degree distributions asymptotically follow a power law \citep{barabasi1999emergence}, 
or unbiased, maximum entropy graph models, with respect to some given constrains \citep{park2004statistical}.
Here we choose a Bernoulli (Erd\"os-R\'enyi) random graph model \citep{solomonoff1951connectivity, erdosRenyi}, for which each link 
is defined via an independent Bernoulli variable with some fixed, global probability $p \in [0,1]$, which is a hyperparameter of our model.
    
 \textbf{Prior over random walks}. The probability $p(\*z_{1:L} | \*A)$ of a random walk over the nodes of $\cal G$ can generally be written as
\begin{multline}
    p(\*z_{1:L}|\*A) = p(\*z_1) \prod^{L}_{i=2} p(\*z_{i} | \*z_{i-1}, \*A) \\ = \left( \prod_{m=1}^K \rho_m^{z_1^m} \right) \prod^{L}_{i=2} \, \left( \prod^K_{j=1} \prod^K_{k=1} P_{k, j}^{z^{k}_i z^{j}_{i-1}} \right),
    \label{eq:prior_rw}
\end{multline}
where $p(\*z_1)$ labels the probability of selecting $j_1$ as the starting point of the walk,
and it is given by (the pdf of) a categorical distribution over the nodes of ${\cal G}$, with class probabilities $\{\rho_i \}_{i=1}^K$.
Similarly $p(\*z_{i} | \*z_{i-1}, \*A)$ labels the conditional probability of jumping from $j_{i-1}$ to $j_i$, which we define in terms of a $K \times K$ transition probability matrix 
$\*P$. 
Now, 
to allow for biased random walks, 
let each node $k$ on $\cal G$ be given a positive weight $f_k$, so that the probability of jumping from $j$ to $k$ is proportional to $f_k \, A_{kj}$.
We then write the transition probability matrix as
\begin{equation}
    P_{k,j} = \frac{f_k \, A_{kj}}{\sum^K_{i=1}  f_i \, A_{ij}},
    \label{eq:transition_matrix_prior}
\end{equation}
so that the motion of the random walker is biased according to the node weights $f_k$.
These weights should be understood as encoding aspects of the diffusion dynamics that are independent of the topology of the graph \citep{gomez2008entropy, lambiotte2011flow}.
Three comments are in order:
first, note that one can also \textit{train} the prior over walks by making the vectors $\+\rho$ and $ \*f$ learnable.
Second, setting the node weights $\*f = \mathbb{I}$ and the class probabilities $\+\rho = \frac{1}{K} \mathbb{I}$, with $\mathbb{I}$ the $K$-dimensional vector of ones, yields a \textit{uniform} random walk over $\cal G$,
i.e. a process in which the walker has equal probability of jumping to any of its neighbors.
Third, one can also allow for \textit{inhomogeneous} random walks in which the probability matrix changes at each step of the random walk. 
Such processes can be parameterized with a sequence of weights $\*f^{[1]}, \*f^{[2]}, \dots, \*f^{[L-1]}$.

\textbf{Decoder and likelihood}. Just as in Eq.~\ref{eq:language_model} we define the joint probability over word sequences as a product of conditional probabilities, each of which is now conditioned on the schema $\*e_{j_1:j_L}$ too. 
Accordingly, the class probabilities of the $i$th conditional are now given by $\+\pi_i = \text{softmax}(  \* W \cdot \*h^{\tiny \text{dec}}_{\theta}(\*x_{<i}, \*e_{j_1:j_L}) )$,
with $\*W \in \mathbb{R}^{V \times D}$ trainable and $\*h^{\tiny \text{dec}}_{\theta}$ a deep neural network model.
We let $\*h^{\tiny \text{dec}}_{\theta}$ be a \textit{pretrained} GPT-2 language model, and modify it to also process the schema $\*e_{j_1:j_L}$, 
but remark that any other model for sequence processing (as e.g. a recurrent neural net) could be used instead.
A bit more in detail, to condition GPT-2 on $\*e_{j_1:j_L}$, without perturbing its optimized weights too much, we use the \textit{pseudo-self}-attention (PSA) mechanism introduced by \cite{ziegler2019encoder}.
In a nutshell, this mechanism augments the key and value matrices of GPT-2 in their first $L$ rows with projections of $\*e_{j_1:j_L}$.
Figure~\ref{fig:model} shows an illustration of the complete decoder model, including the PSA mechanism.
Please check Appendix \ref{app:PSA} for the explicit equations of the latter. 

\subsection{Inference Model}

The generative model we presented above is hierarchical.
The random graph is shared across all sentences and thus constitutes a global latent object.
The random walks, in contrast, are local random variables.
Our task is to infer the schema and graph posterior distributions that best describe the collection of word sequences in our datasets.
To do this, we approximate the true posterior distribution of these variables with a variational posterior of the form
\begin{equation}
    q_{\phi}(\*z_{1:L}, \*A| \*x_{1:T}) = q_{\phi}(\*z_{1:L}| \*x_{1:T}, \*A) q_{\phi}(\*A),
    \label{eq:full_posterior}
\end{equation}
where $\phi$ labels the set of trainable parameters. 
We model $q_{\phi}(\*A)$, the posterior over the graph, assigning again Bernoulli variables to its links, 
but we let the probability of observing each link depend on the global symbols. 
That is, we model the link probabilities as $p_{\phi}(\*e_i, \*e_j) = \text{sigmoid}(g_{\phi}(\*e_i, \*e_j))$, for all $\*e_i \in {\cal E}$,
with $g_{\phi}: {\cal E} \times {\cal E} \rightarrow \mathbb{R}$ a deep neural network.

Likewise we model $q_{\phi}(\*z_{1:L}| \*x_{1:T}, \*A)$, the posterior probability over random walks on the graph, analog to Eq.~\ref{eq:prior_rw}.
This time, however, instead of having a single transition probability matrix, we have a sequence of them,
thereby allowing the posterior to capture inhomogeneous random walks.
We therefore model the class probabilities over the starting point of the random walks as $\+\rho(\*x_{1:T}, \phi) = \text{softmax}(\*h^{\tiny \text{enc}}_1)$, whereas the sequence of transition matrices is given by
\begin{eqnarray}
Q_{k,j}^{[i]}(\*x_{1:T}, \*A, \phi)  =  \frac{f^{[i]}_k(\*x_{1:T}, \phi) \, A_{kj}}{\sum_m  f^{[i]}_m(\*x_{1:T}, \phi) \, A_{mj}},  \label{eq:transition_prob_matrix_q}
\end{eqnarray}
with the sequence of node weights 
$\*f^{[1]}, \*f^{[2]} $ $\dots, \*f^{[L-1]} = \exp(\*h^{\tiny \text{enc}}_{2:L})$.
Now, the set 
$\*h^{\tiny \text{enc}}_1, \*h^{\tiny \text{enc}}_2, $ $\dots, \*h^{\tiny \text{enc}}_L \in \mathbb{R}^K$ 
is the sequence of outputs of a deep neural network model $\*h^{\tiny \text{enc}}_{\phi}(\*x_{1:T})$ processing the input sequence of $T$ words.
The model $\*h^{\tiny \text{enc}}_{\phi}(\*x_{1:T})$ must then map a sequence of $T$ vectors to a sequence of $L$ vectors.
We define $\*h^{\tiny \text{enc}}_{\phi}$ by a \textit{pretrained} BERT model \citep{devlin2018bert}, followed by a single Transformer block, randomly initialized.
The Transformer block processes the $T$ ($D$-dimensional) outputs from BERT as keys and values, together with a set of $L$ learnable vectors $\*q_{1:L}$ as queries. 
The right-hand side of Figure~\ref{fig:model} illustrates the complete encoder architecture. 
For completeness, we also give the explicit and complete expressions of both posterior graph and random walk models in Appendix \ref{sec:posterior-full}.

\subsection{Training Objective}

To optimize the parameter sets $\{\theta, \phi\}$ of our latent variable model we would, as usual, maximize a variational lower bound on the logarithm of the marginal likelihood $p_{\theta}(\*x_{1:T})$ \citep{bishop2006pattern}. 
It is, however, well known that VAE models tend to encounter problems learning representations encoding information about the data -- the so-called posterior collapse problem -- especially when dealing with natural language \citep{bowman2015generating}. 
To solve this issue practitioners resort to maximizing the variational lower bound, together with the mutual information between data and representations \citep{zhao-etal-2018-unsupervised, fang-etal-2019-implicit, Zhao_Song_Ermon_2019}.
We follow this same route here and refer the reader to
Appendix \ref{app:training_objective} for the derivation of our objective function. 
The training algorithm is presented in Appendix~\ref{sec:algorithm} and training times in Appendix~\ref{app:training_times}.
Source code to reproduce all experiments is available online.\footnote{Source code: \url{https://github.com/ramsesjsf/HiddenSchemaNetworks}}

\begin{table*}[t!]
\centering
\footnotesize
\begingroup
\setlength{\tabcolsep}{6pt} \renewcommand{\arraystretch}{1.2} \begin{tabular}{cccccc} 
    \hline
     Graph ${\cal G^*} $ & ROC AUC & $||{\cal G^*} - {\cal G}||_F$ & $||{\cal G^{\tiny \text{rand}}} - {\cal G}||_F$ & N. edges(${\cal G} $) & N. edges(${\cal G^*}) $   \\
    \hline 
    
    \hline
    \texttt{Barabasi}   & 0.989 $\pm$ 0.001   & 17 $\pm$ 2    & 26 $\pm$ 1 & 1360 $\pm$ 104 & 291\\   

    \texttt{Erdos} &  0.94 $\pm$ 0.06    & 36.8 $\pm$ 0.8 & 44 $\pm$ 2 & 3131 $\pm$ 156 & 2092\\     
    \hline
\end{tabular} \endgroup
\caption{Results on ground-truth random graphs inference. $\cal G^*$ ($\cal G$) labels ground-truth (discovered) graph. $||\cdot ||_F$ labels Frobenius norm. Error bars are computed from 10 random model initializations.}
\label{tab:synthetic}
\end{table*}
 \label{sec:model}

\section{Inferring Ground-Truth Graphs}
Before testing the behavior of our methodology on natural language data, we evaluate the ability of the model to infer hidden graph structures from sequential data in a controlled experiment.
To this end, we define a synthetic language model with an underlying, ground-truth graph ${\cal G^*}$ as follows:
Given a graph ${\cal G^*}$ with $K$ nodes, and a vocabulary of random tokens ${\cal V}$ of size $V$, we assign one random bag of tokens (i.e. one pdf over ${\cal V}$) to each node of the graph.
Let the $K$ random bags be the set ${\cal E}$ of $K$ symbols $\{\*e_1, \*e_2, \dots, \*e_K\}$ of the synthetic language model.
We then sample $N$ uniform random walks of length $L$ over ${\cal G^*}$, and sample one random token from each symbol (i.e. from each random bag) along the walks. 
The result is a set of random token sequences of the same length as that of the random walks.

 \textbf{A simple proof-of-concept}. We consider a problem in which the set of symbols $\cal E$ is known, so that the ground-truth graph ${\cal G^*}$ has a fixed labeling. 
This setting will allow for simple comparison between ${\cal G^*}$ and our inferred graphs.
Following the procedure above we generate two datasets from two ground-truth graphs with different topologies. One sampled from the Barab\'asi-Albert model \citep{barabasi1999emergence}, the other from the Erd\"os-R\'enyi model \citep{erdosRenyi}. 
We set both graphs to have $K=100$ symbols, and the token sequences to have length $L=10$.
Given these sets of random token sequences, the task is to infer the hidden, ground-truth graphs ${\cal G^*}$.
To solve it we first replace BERT in Fig.~\ref{fig:model} with a 2-block Transformer encoder \citep{vaswani2017attention}
and note that, since the symbols are known, the likelihood of the model is simply given by $\prod_{i=1}^L \*e_{j_i}$ where, as before, $j_i$ denotes the index of the non-zero component of $\*z_i$.
We then train this simplified model by maximizing the HSN objective function. 
Further details about the graph model parameters, the dataset generation procedure and statistics, training procedure, and model sizes and hyperparameters can be found in Appendix \ref{app:synthetic}. 
The synthetic datasets are available in the source code too.
 
 \textbf{Results}. Table~\ref{tab:synthetic} shows our results for our two synthetic datasets. 
Specifically, we compute the Area Under the Receiver Operating Characteristic Curve (ROC AUC) of our model $q_{\phi}(\*A)$ with respect to ${\cal G^*}$, and the Frobenius norm between $q_{\phi}$ and two graphs: the ground-truth one ${\cal G^*}$, and a second random graph ${\cal G^{\tiny \text{rand}}}$ sampled from the same random graph model as ${\cal G^*}$. 
We train ten (10) models in total and display the mean and standard deviation of our results. 
We also use a different ${\cal G^{\tiny \text{rand}}}$ for each calculation run.
The first metric shows that $q_{\phi}(\*A)$ correctly predicts the edges of ${\cal G^*}$, whereas the other two metrics show that ${\cal G} \sim q_{\phi}(\*A)$ is closer to  ${\cal G^*}$, than to any other random graph sampled from the same distribution.
The last two columns in Table~\ref{tab:synthetic} show however that $q_{\phi}(\*A)$ tends to generate denser graphs as compared to the target.
This overproduction of edges can be understood as being caused by the  prior graph model, which is chosen to be an Erd\"os-R\'enyi model (see Section \ref{sec:gen_model}).  
The expected number of edges of this model is $\binom{K}{2} p$, where $p$ is the prior edge probability. 
In our experiments we have $K=100$, and choose $p = 0.2$ ($0.6$) when training the model on the Barabasi (Erd\"os) dataset (see Appendix~\ref{app:synthetic}). The average number of edges of the prior graph model is therefore 990 (2970) for the Barabasi (Erd\"os) dataset. 
During training, the posterior model tries to fit the data, while simultaneously being close (in distribution) to the prior, which explains the overproduction of edges. Using different priors, like e.g.~a sparse graph model, would reduce the average number of edges of the posterior graph distribution. See also Table~\ref{tab:results_synthetic_sup_mat} in Appendix~\ref{app:synthetic}, where we report results from HSN with smaller prior edge probabilities.
 
\begin{figure*}[t!]
\footnotesize
\newfloatcommand{capbtabbox}{table}[][\FBwidth]
\begin{floatrow}
    \capbtabbox{

\begin{tabular}{ l  cc   cc   cc  } 
\hline
 & \multicolumn{2}{c}{\texttt{\footnotesize PTB}} &  \multicolumn{2}{c}{\texttt{\footnotesize YAHOO}} &  \multicolumn{2}{c}{\texttt{\footnotesize YELP}} \\ 
\footnotesize Model                      & PPL & MI  & PPL & MI  & PPL & MI \\
\hline
\hline
\footnotesize GPT2   & \footnotesize 24.23  & -- & \footnotesize 22.00  & -- & \footnotesize  23.40  & -- \\
\footnotesize $\text{iVAE}_{\tiny \text{MI}}$  & \footnotesize 53.44 & \footnotesize 12.50 & \footnotesize  47.93 & \footnotesize 10.70 & \footnotesize  36.88 & \footnotesize 11.00 \\
\footnotesize $\text{Optimus}_{ \tiny \text{A}}$          & \footnotesize 23.58 & \footnotesize 3.78  & \footnotesize 22.34 & \footnotesize 5.34 & \footnotesize  21.99  & \footnotesize 2.54\\
\footnotesize $\text{Optimus}_{ \tiny \text{B}}$          & \footnotesize 35.53 & \footnotesize 8.18  & \footnotesize 29.92 & \footnotesize 9.18 & \footnotesize  24.59  & \footnotesize 9.13\\
\hline
\footnotesize $\text{HSN}_{ \tiny \text{(50, 5)}}$  & \footnotesize 22.47 & \footnotesize 9.50  & \footnotesize \textbf{20.99} & \footnotesize 10.42 & \footnotesize  \textbf{19.72} & \footnotesize 10.04 \\
\footnotesize $\text{HSN}_{ \tiny \text{(50, 20)}}$        & \footnotesize 30.38 & \footnotesize \textbf{26.06}  & \footnotesize 22.84 & \footnotesize \textbf{22.81} & \footnotesize  21.60 & \footnotesize \textbf{24.93} \\
\footnotesize $\text{HSN}_{ \tiny \text{(100, 5)}}$ & \footnotesize \textbf{20.25} & \footnotesize 9.30  & \footnotesize 21.01 & \footnotesize 11.21 & \footnotesize  19.82 & \footnotesize 10.20 \\
\footnotesize $\text{HSN}_{ \tiny \text{(100,20)}}$  & \footnotesize 25.48 & \footnotesize 23.77  & \footnotesize 21.98 & \footnotesize 16.13 & \footnotesize  21.13 & \footnotesize 18.85 \\

\hline
\end{tabular}

     }{
        \caption{Perplexity (PPL) (lower is better) and mutual information (MI). 
$\text{Optimus}_{ \tiny \text{A, B}}$ label models with best PPL and MI, respectively (with $\lambda=0.05, 1$) \citep{li-etal-2020-optimus} .
GPT2 results are taken from \citet{li-etal-2020-optimus}.
$\text{iVAE}_{\tiny \text{MI}}$ was taken from \citet{fang-etal-2019-implicit}. We sampled 100 (10) random walks (graphs) to estimate the PPL. End-of-sequence tokens are kept during evaluation.}
        \label{tab:results_lm}
    }
    
    \ffigbox[\FBwidth]{
        \includegraphics[width=0.35\textwidth]{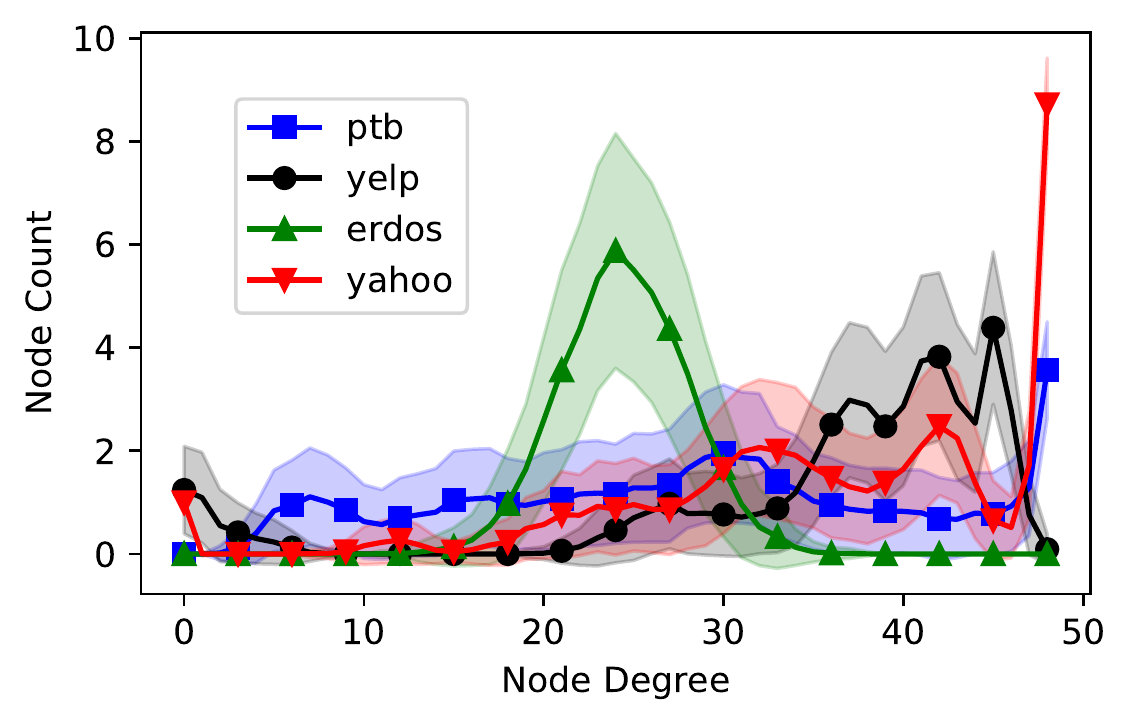}
    }{
        \caption{ Empirical degree distributions of inferred schema networks against that of an Erd\"os-R\'enyi graph with $p=0.5$.
        Results correspond to $\text{HSN}_{ \tiny \text{(50, 5)}}$. The graphs are sampled 500 times. Note that HSN differ from simple random graphs.}
        \label{fig:degree_distribution}
    }
\end{floatrow}
\end{figure*}

 \label{sec:synthetic__}

\section{Natural Language HSN}
In the previous section we demonstrated that HSN can indeed infer hidden graph structure from sequential data in a simple setting.  
We now move on to our main problem: representation learning from natural language. 
We leverage pretrained BERT and GPT-2 language models as encoder and decoder, respectively, which we finetune to discover schema networks encoding natural language datasets.
We consider three widely used public datasets, namely the Penn Treebank (PTB) 
\citep{marcus-etal-1993-building}, Yahoo and Yelp \citep{DilatedCNNVAE} corpora,
and evaluate the quality of the inferred representations via the mutual information (MI) between schemata and data, and the perplexity (PPL) of the model. 
The latter is estimated through Monte Carlo samples.
We compare HSN against a pretrained GPT-2, fine-tuned during a single epoch.
We also compare against two VAE language models: $\text{iVAE}_{\tiny \text{MI}}$ \citep{fang-etal-2019-implicit} and Optimus \citep{li-etal-2020-optimus}. The former implements both encoder and decoder as one-layer LSTMs \citep{10.1162/neco.1997.9.8.1735}. The latter does it via pretrained BERT and GPT-2, just as HSN.
Further details on the definition and computation of the evaluation metrics, as well as on hyperparameters and training procedures can be found in Appendix \ref{app:additional_results}.
In what follows we explore HSNs with $K=\{50, 100\}$ symbols and hidden random walks of $L=\{5, 20\}$ steps.
Let us label these configurations as $\text{HSN}_{ \tiny (K, L)}$.

\begin{figure*}[t!]
    \centering
    \includegraphics[width=\textwidth]{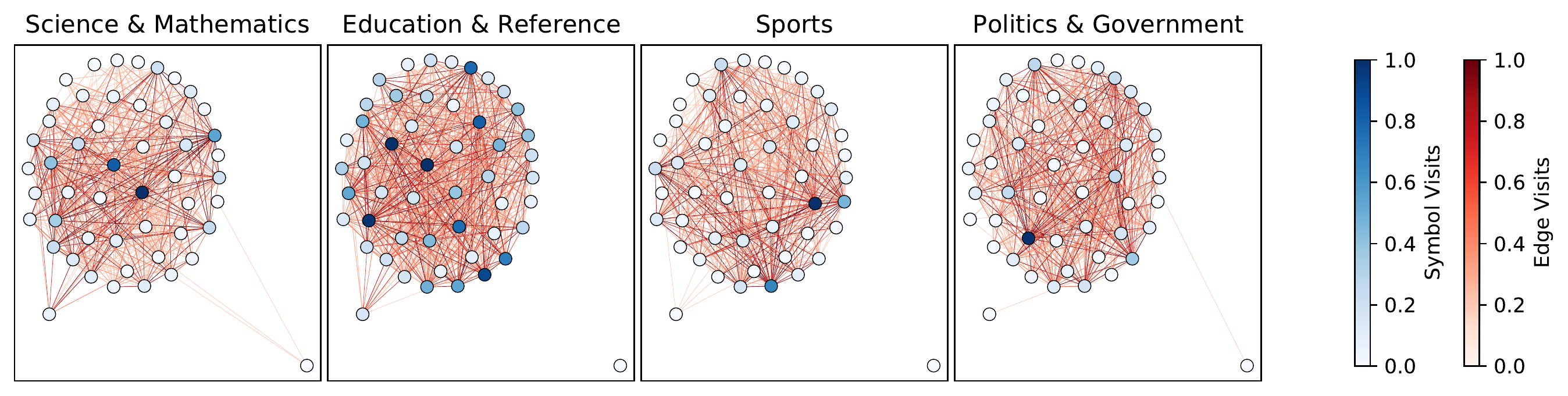}
    \includegraphics[width=\textwidth]{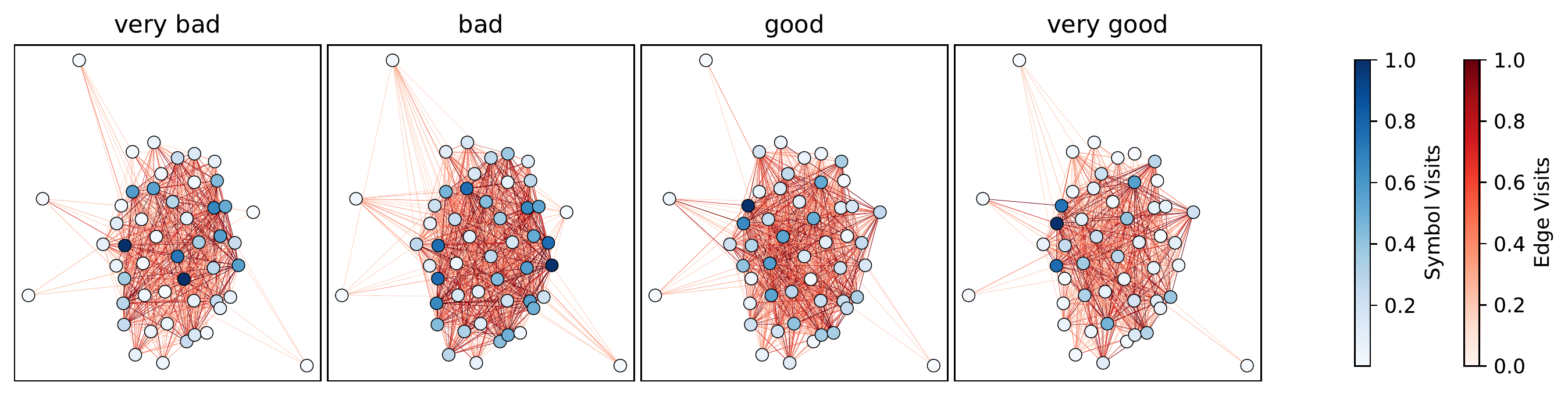}
    \caption{Schema distributions inferred from $\text{HSN}_{ \tiny \text{(50, 5)}}$ for four subsets of the Yahoo (top) and Yelp (bottom) corpora. The node positions in the figure are consistent among labels and were computed using a force-directed embedding of the global graph $\cal G$.
    }
\label{fig:graph_yahoo}
\end{figure*}

\textbf{Results}. As shown in Table~\ref{tab:results_lm},
HSN achieves a much better performance than all baselines under both metrics, which
implies that (i) HSN successfully encodes sentences into symbol sequences, and that (ii) our modified GPT-2 can effectively interpret symbolic representations.
Longer random walks clearly have more capacity and, accordingly, encode more information.
In contrast, shorter random walks perform better wrt.~PPL, which illustrates the usual trade-off between language modeling and representation learning, typical of VAE  \citep{li-etal-2020-optimus}.
We speculate the reason why the structured representations of HSN help with language modeling might be, that they allow for a sequential composition of the sentences' semantic components via reusable units. This feature might yield more efficient encodings than dense representations do. 
HSN also allows to infer (graph) connections not directly seen during training, that account for unseen symbol combinations, and hence (might account for) unseen sentences.
We additionally report in Appendix \ref{app:additional_results} the averaged KL and likelihood values from five HSN, trained with different random initializations, against the baselines.
Now we investigate the structure and semantics of the inferred schema networks.

\textbf{Structure of hidden schema networks}. 
We characterize the structure of the schema networks we inferred in terms of five statistics, namely their diameter, average distance, clustering coefficient, number of connected components and degree distribution.
We define all these and report their computed values, for all HSN configurations and all datasets, in Appendix \ref{app:additional_results}.
One observation we can immediately make about these results is that 
the schema networks discovered in each corpora tend to have smaller average distances, and much larger clustering coefficients, than any random graph of the same size, sampled from a maximum-entropy ($p=0.5$) Erd\"os-R\'enyi model.
Interestingly enough, the combination of these two features defines the so-called \textit{small-world structure} \citep{watts1998collective}.
Now, intuitively speaking, a larger clustering coefficient implies that a random walker starting from a given node ``$k$'' will have a larger number of paths bringing it back to ``$k$''.
In such an scenario, random walkers tend to cluster in neighborhoods around their starting point -- a property that could help encode different semantic aspects in different regions of ${\cal G}$.
Similarly, one could also expect schemata composed of repeated symbols.
We close this subsection with Figure \ref{fig:degree_distribution}, which displays the difference between the degree distributions of the schema networks and that of purely random graphs.

\textbf{Schemata and semantics}. 
Taking advantage of the labels available to both Yahoo and Yelp, we display in Figure~\ref{fig:graph_yahoo} the schema distributions over the schema networks for 4 subsets of these datasets.
Similar plots for all subsets of both corpora, extracted with all HSN configurations, can be found in Appendix \ref{app:additional_results}. 
Note how the ``hot'' symbols per category reside on different regions of the graphs -- as suspected already from the large clustering coefficient of ${\cal G}$ -- 
and yet, the ``Science \& Math'' schemata (both nodes and edges) of Yahoo are \textit{closer} to the ``Education \& Reference'' schemata than to the ``Sport'' schemata,
where closer nodes in the figure indicate well-connected nodes in the underlying graph ${\cal G}$.
These findings (qualitatively) indicate that the schemata indeed encode semantic notions of their corpora.
A similar picture holds for Yelp.
We also introduce and explore ``schema interpolations'' in Appendix~\ref{subsec:app_addional_results}, and observe that the translation procedure we employ successfully interpolates between categories of both Yelp and Yahoo.
These experiments hint too at the different semantic notions encoded by HSN.

To get a more quantitative evaluation of the semantics encoded in the schema networks, one can also investigate how the distribution of attention weights, in the HSN decoder, and wrt.~the symbols changes, as the decoder processes specific words. 
We exemplify this idea in Appendix~\ref{sec:attention_to_symbols}.

 \label{sec:LM}

\section{Commonsense Reasoning Generation}
\begin{table*}[!htbp]
\centering
\footnotesize
\begingroup
\setlength{\tabcolsep}{6pt} \renewcommand{\arraystretch}{1.2}

\begin{tabular}{cccccc} 
\hline
  & $\text{HSN}_{ \tiny \text{(50, 20)}}$ & $\text{HSN}_{ \tiny \text{(50, 20)}}^{\tiny \text{AR}}$ & COMET\tiny(GPT2)  & COMET\tiny(GPT2-XL) & COMET \tiny(BART) \\
  \hline
 BLEU-2 & \textbf{0.462} & 0.129 & 0.225 & 0.300 & 0.330  \\
 BERT Score & \textbf{0.694} & 0.374 & 0.385 & 0.638 & 0.650 \\
\hline
\end{tabular} \endgroup
\caption{Object generation quality. COMET{\tiny(GPT2-XL)} and COMET {\tiny(BART)} results were extracted from \cite{hwang2020comet}. COMET{\tiny(GPT2)} was computed by us. All models use greedy decoding for \textit{all} text prefixes in the dataset.}
\label{tab:atomic}
\end{table*}

In the previous section we empirically showed that HSN is able to infer schema networks from natural language datasets, whose symbols encode different notions of semantics. 
Next, we explore training ``reasoning'' models on the schema networks inferred from commonsense knowledge databases, and using these secondary models to sample novel (reasoning) paths and enhance the performance of LPLM on commonsense \textit{If-Then} reasoning tasks.
Let us start by defining this task.

It has been recently proposed that LPLMs fine-tuned on (natural language) knowledge graph (KG) tuples, can express their encoded knowledge through language generation, thereby providing commonsense knowledge on demand \citep{bosselut2019comet, hwang2020comet}. 
Consider a KG of natural language tuples of the form ($\*s$, $\*r$, $\*o$), where $\*s=(\*x^s_1, \dots, \*x^s_{|\*s|})$ labels the \textit{phrase subject} of the tuple, $\*r=\*x^r$ is the \textit{relation token} and $\*o=(\*x^o_1, \dots, \*x^o_{|\*o|})$ is the \textit{phrase object} of the tuple.
In this setting, commonsense reasoning consists in generating novel objects $\*o$, given $\*s$ and $\*r$.

The COMET framework of \citet{bosselut2019comet} finetunes GPT-like models by maximizing the likelihood of the object, conditioned on the sequence $(\*s, \*r)=(\*x^s_1, \dots, \*x^s_{|\*s|}, \*x^r)$.
We instead approach the problem of commonsense knowledge generation with a three-step process:
(I) we auto-encode the KG tuples using HSN so that the first $\frac{L}{2}$ symbols of the schemata encode $(\*s, \*r)$ only, while their last $\frac{L}{2}$ symbols encode the complete ($\*s$, $\*r$, $\*o$) tuple;
(II) we train an autoregressive (AR) 
``reasoning''
model, that takes $(\*s, \*r)$ and $\*z_{1:\frac{L}{2}}$ (the first $\frac{L}{2}$ schema symbols inferred in step I) as inputs, and learns to generate the second half of the schema $\*z_{\frac{L}{2}+1:L}$ (the half that encodes the object); 
(III) given unseen $(\*s, \*r)$ pairs and their associated first half-schemata $\*z_{1:\frac{L}{2}}$, we sample novel, second half-schemata $\*z_{\frac{L}{2}+1:L}$ from the trained AR model, and use them to condition the HSN decoder and generate novel objects.

To do step I we write the random walk posterior distribution $q_{\phi}(\*z_{1:L}|\*s, \*r, \*o)$ of HSN in the product form
$q_{\phi}(\*z_{1:\frac{L}{2}}| \*s, \*r) q_{\phi}(\*z_{\frac{L}{2}+1:L}| \*z_{\frac{L}{2}}, \*s, \*r, \*o)$,
where we omitted the explicit dependence on $\*A$, in all terms, for clarity. 
Both terms in the product are modeled with the same HSN encoder architecture as before (see Fig.~\ref{fig:model}), but share a single, pretrained BERT model.
To do step II we introduce an AR ``reasoning'' model $p_{\theta}(\*z_{\frac{L}{2} + 1:L}| \*z_{1:\frac{L}{2}}, \*s, \*r)$, that processes the first half-schema $\*z_{1:\frac{L}{2}}$ via a recurrent neural net, and the pair $(\*s, \*r)$ via a second, pretrained BERT model. 
This AR model is trained to generate the second half-schema on samples from $q_{\phi}(\*z_{\frac{L}{2}+1:L}| \*z_{\frac{L}{2}}, \*s, \*r, \*o)$.
Finally, we do step III by concatenating the first half-schema, sampled from the posterior model, with the second half-schema, sampled from the AR model.
The novel reasoning path is used to condition the HSN decoder to generate novel objects. 
See Appendix \ref{app:commonsense} for additional details.

For this preliminary study we focus on the ATOMIC dataset \citep{sap2019atomic},
evaluate the quality of the generated objects with both, BLEU-2 \citep{papineni-etal-2002-bleu} and BERT Score \citep{Zhang*2020BERTScore} metrics,
and compare against GPT-2, GPT-2-XL and BART, all trained within the COMET framework \citep{hwang2020comet}. 
We also fix the number of symbols $K = 100$, and the random walk length $L=20$.

\textbf{Results.} We find that our best AR models only learn about $58\%$ of the second half-schema. That is, about half of the symbols encoding the complete ($\*s$, $\*r$, $\*o$) tuple.
Table \ref{tab:atomic} shows the scores we obtain when conditioning the decoder on the schemata sampled from the posterior (which sees the target object), and from the AR model (which does not sees the object): $\text{HSN}_{ \tiny \text{(50, 20)}}$ and $\text{HSN}_{ \tiny \text{(50, 20)}}^{\tiny \text{AR}}$, respectively.
That $\text{HSN}_{ \tiny \text{(50, 20)}}$ outperforms all baselines is expected, for HSN can successfully encode its inputs, as we empirically showed in Section~\ref{sec:LM} above.
We find interesting that $\text{HSN}_{ \tiny \text{(50, 20)}}^{\tiny \text{AR}}$ is comparable to COMET{\tiny(GPT2)} in BERT Score, despite leveraging an imperfect AR model, which only reproduces about $58\%$ of the correct encoding.
The scores achieved by $\text{HSN}_{ \tiny \text{(50, 20)}}$ imply that improving upon the AR model should increase the performance of $\text{HSN}_{ \tiny \text{(50, 20)}}^{\tiny \text{AR}}$, even beyond that of COMET{\tiny(BART)}.
We speculate that the AR ``reasoning'' model struggles to infer the schema distribution because of the complexity of the latter. 
This complexity is direct consequence of the object-encoding scheme we use (step I above), for which e.g.~missing one single symbol can lead to random walks completely off.
We shall investigate different encoding schemes in future work.
 \label{sec:commonsense}

\section{Conclusion}

We have introduced a novel representation learning algorithm for natural language that (i) enforces, via inductive biases, relational structures which allow for compositionality onto the output representations of LPLM,
and (ii) allows for symbolic reasoning via random walk processes on the discovered graphs.
Future work shall investigate replacing both encoding and decoding models with more expressive LPLM,  
and explore training different ``reasoning'' models and encoding schemes, on natural language understanding tasks.

\section*{Limitations}

Although our methodology is agnostic to the specific sequence-processing models used as encoder and decoder, its main purpose is to be used together with LPLM. 
This clearly limits its use to institutions and users with access to computing facilities able to handle such models.
Indeed, although the encoder of HSN can successfully be trained while keeping the weights of BERT (or any other LPLM) frozen, a feature which reduces training time, the GPT-based decoder does need to be fine-tuned to learn how to interpret the inferred symbols. 
This last points further limits the use of our methodology to ``not-so-large'' decoder models. That being said, fine-tuning HSN decoders via LoRA \cite{hu2022lora} is an exciting option that can possibly solve (or help with) this last issue. We shall explore using LoRA with HSN in the near future.

\section*{Ethics Statement}
We do not find any ethical concerns either with the model or the datasets we used (which are publicly available). 

\section*{Acknowledgements}
This research has been funded by the Federal Ministry of Education and Research of Germany and the state of North-Rhine Westphalia as part of the Lamarr-Institute for Machine Learning and Artificial Intelligence, LAMARR22B. 
Part of this work has been funded by the Vienna Science and Technology Fund (WWTF) project ICT22-059. César Ojeda has been partially supported by Deutsche Forschungsgemeinschaft (DFG) Project-ID 318763901 - SFB1294 as well as
Project-ID 235221301 - SFB 1114. Part of the work was also funded by the BIFOLD-Berlin Institute for the Foundations of Learning and Data (ref. 01IS18025A and ref 01IS18037A). C.O. would like to thank Manfred Opper for inspiring discussions, and Sarah Elena M\"uller for relevant literature.

\bibliography{bibliography}
\bibliographystyle{acl_natbib}

\clearpage
\section*{\Large Appendix}
\appendix
\section{Pseudo-Self Attention Mechanism Revisited}
\label{app:PSA}

The attention mechanism of the original Transformers \citep{vaswani2017attention} is defined as 
\begin{multline}
    \text{Attention}(\*Q, \*K, \*V) = \\ \text{softmax} \left( D^{-\frac{1}{2}} \, \*Q \cdot \*K^T\right) \cdot \*V,
    \label{eq:attention}
\end{multline}
where $\*Q$, $\*K$ and $\*V \in \mathbb{R}^{T\times D}$ are sets of queries, keys and values, respectively, given by a sequence of $T$, $D$-dimensional vectors, packed into matrices. In practice, these queries, keys and values are projected many times with different learnable, linear maps. The $\text{Attention}(\cdot)$ operation (Eq.~\ref{eq:attention}) is performed on these different projections in parallel, whose outputs are then concatenated and projected once more with a final, linear map. The complete operation is known as Multi-head Attention \citep{vaswani2017attention}, and we use this notation in Fig. \ref{fig:model} of the main text.

Now, the question is how to condition GPT-2 on the schema $\*e_{j_1:j_L}$.
Given a sequence of input representations $\*u_{1:T}$, the \textit{self}-attention mechanism in GPT-2 is obtained by choosing $\*Q = \*u_{1:T} \cdot \*W_Q, \*K= \*u_{1:T} \cdot \*W_K$ and $\*V = \*u_{1:T} \cdot \*W_V$, all in $\mathbb{R}^{T \times D}$, with $\*W_Q, \*W_K$ and $\*W_V \in \mathbb{R}^{D\times D}$ pretrained matrices. 
We leverage a pseudo-self attention (PSA) mechanism \citep{ziegler2019encoder} that augments the key and value matrices in their first $L$ rows, with projections of $\*e_{j_1:j_L}$ so that
\begin{align}
 \tilde{\*K} &= \begin{pmatrix}  \*e_{j_1:j_L} \cdot \*W^e_K  + \*p_{\tiny \text{enc}} \\  \*K \end{pmatrix}, \, \nonumber \\ 
 \tilde{\*V} &= \begin{pmatrix}  \*e_{j_1:j_L} \cdot \*W^e_V + \*p_{\tiny \text{enc}} \\  \*V \end{pmatrix},
 \label{eq:psa}
\end{align}
both in $\mathbb{R}^{(L + T)\times D}$,
where $\*p_{\tiny \text{enc}}$ is a positional encoding, just as the one used in the original Transformer implementation \citep{vaswani2017attention}.
The latter informs GPT-2 about the ordering of the symbols in the schema, as selected by the random walk process.
PSA is then simply given by Eq.~\ref{eq:attention} with the keys and values replaced with the augmented ones, $\tilde{\*K}$ and $\tilde{\*V}$.
The $\*W^e_K, \*W^e_V$ here are randomly initialized, learnable parameters mapping the schemata onto the decoder self-attention, $D$-dimensional space, and we have as many of them as layers in GPT-2.
Therefore this mechanism allows GPT-2 to attend to the projected schema at each of its layers, with a minimal addition of untrained parameters \citep{ziegler2019encoder}.

\section{Hidden Schema Networks Algorithm}
\label{sec:algorithm}

\RestyleAlgo{ruled}
\SetKwComment{Comment}{/* }{ */}

\begin{algorithm}[h!]
\caption{HSN Training ($\phi$, $\psi$)}\label{alg:model_training}
\ForEach {\text{minibatch} $\*x_{1:T} \sim p(\mathcal{D})$}{
 \BlankLine \BlankLine
    \textbf{(1) Sample schema network from posterior graph model:} 
    \begin{eqnarray}
    \*A & \sim & q_\phi(\*A), \nonumber
    \end{eqnarray}
    
\textbf{(2) Compute parameters of posterior random walk model:}
\begin{eqnarray*}
\*h_1^{\tiny \text{enc}}, \*h_2^{\tiny \text{enc}} \dots, \*h_L^{\tiny \text{enc}} & = & \*h_{\phi}^{\tiny \text{enc}}(\*x_{1:T}), \nonumber \\
\+\rho(\phi) & = &\text{softmax}(\*h^{\tiny \text{enc}}_1), \nonumber \\ 
Q_{k,j}^{[i]}(\phi) & = & \frac{f^{[i]}_k(\phi) \, A_{kj}}{\sum_m  f^{[i]}_m(\phi) \, A_{mj}}, \, \, \, \nonumber \\
\end{eqnarray*}
with  $\*f^{[1]}, \dots, \*f^{[L-1]} = \exp(\*h^{\tiny \text{enc}}_{2:L})$
\vspace{0.5em}

\textbf{(3) Compute parameters of prior random walk model:}
    \begin{equation*}
    P_{k,j} = \frac{f_k \, A_{kj}}{\sum^K_{i=1}  f_i \, A_{ij}}
    \end{equation*}
    
\textbf{(4) Sample random walks from posterior distribution:}
\begin{equation*}
    \*z_{1:L} \sim q_{\phi}(\*z_{1:L}| \*x_{1:T}, \*A)
\end{equation*}

\textbf{(5) Decode sentence:}
\vspace{0.5em}

    \For{$i=0$ \KwTo $T-1$}{
$\*x_i \sim p_{\theta}(\*x_{i} | \*x_{<i}, \*e_{j_1:j_L}),$
        
        $\+\pi_i = \text{softmax}(  \* W \cdot \*h^{\tiny \text{dec}}_{\theta}(\*x_{<i}, \*e_{j_1:j_L}) ) $
}
    \vspace{0.5em}
  
\textbf{(6) Compute loss and back-propagate:}
\begin{multline*}
    \mathcal{L}[\theta, \phi] = \frac{1}{N} \, \sum_{n=1}^N    \mathbb{E}_{q_{\phi}(\*z^{\phantom{(n)}}_{1:L}|\*x^{(n)}_{1:T}, \, \*A) q_{\phi}(\*A)} \\
\Big\{ \log p_{\theta}(\*x^{(n)}_{1:T} | \*z^{\phantom{(n)}}_{1:L}) \Big\} \\
- \mathbb{E}_{q_{\phi}(\*A)} \text{KL}\Big[q^*_{\phi}(\*z_{1:L}| \*A); p(\*z_{1:L}| \*A) \Big] \\
- \text{KL}[q_{\phi}(\*A);p(\*A)]
\end{multline*}
}
\end{algorithm}

\section{Inference model: Full Equations}
\label{sec:posterior-full}

\subsection{Posterior over (global) graph}

We model the posterior over the graph assigning Bernoulli variables to its links, 
but we let the probability of observing each link depend on the global symbols
\begin{multline}
\small
    q_{\phi}(\*A) = \prod_{i,j} p_{\phi}(\*e_i, \*e_j)^{a_{ij}} \left(1-p_{\phi}(\*e_i, \*e_j) \right)^{1-a_{ij}} \, \\ 
    \text{where} \, \, \, 
    p_{\phi}(\*e_i, \*e_j) = \text{sigmoid}(g_{\phi}(\*e_i, \*e_j)),
     \nonumber
\end{multline}
with $g_{\phi}: {\cal E} \times {\cal E} \rightarrow \mathbb{R}$ a deep neural network, and $p_{\phi}(\*e_i, \*e_j) \in [0, 1]$, for all $\*e_i \in {\cal E}$, the link probabilities.
Our reasoning here is that the network $g_{\phi}$ should infer graphs connecting symbols which are semantically related via the encoded sentences.

\subsection{Posterior over random walks (encoder model)}

We model the posterior probability over random walks on $\cal G$ as
\begin{multline}
\small
    q_{\phi}(\*z_{1:L}| \*x_{1:T}, \*A) = \left( \prod_{i=1}^K \rho_i(\*x_{1:T}, \phi)^{z_1^i} \right) \\
    \cdot \prod^{L}_{i=2} \, \left( \prod^K_{j=1} \prod^K_{k=1} \left(Q_{k, j}^{[i-1]} (\*x_{1:T}, \*A, \phi) \right)^{z^k_{i} z^j_{i-1}} \right), 
     \label{eq:posterior_rw_app}
\end{multline}
where instead of having a single transition probability matrix, we have a sequence of them,
thereby allowing the posterior to capture inhomogeneous random walks.
Note that we could have also chosen a mean-field decomposition along the steps of the random walk, simply by either ignoring the dependency on the graph, or making the graph fully connected (see Appendix \ref{subsec:MF}). 
Going back to Eq.~\ref{eq:posterior_rw_app}, we model the probabilities over the starting point of the random walks and the transition matrices as follows
\begin{eqnarray}
\small
\+\rho(\*x_{1:T}, \phi) & = &\text{softmax}(\*h^{\tiny \text{enc}}_1),
\label{eq:starting_point_rw_q} \nonumber \\ 
Q_{k,j}^{[i]}(\*x_{1:T}, \*A, \phi) & = & \frac{f^{[i]}_k(\*x_{1:T}, \phi) \, A_{kj}}{\sum_m  f^{[i]}_m(\*x_{1:T}, \phi) \, A_{mj}}, \, \, \, \nonumber\\  \text{with}  \, \, \,  \*f^{[1]}, \dots, \*f^{[L-1]} & = & \exp(\*h^{\tiny \text{enc}}_{2:L}), \nonumber \label{eq:transition_prob_matrix_q_app}
\end{eqnarray}
where $\*h^{\tiny \text{enc}}_1, \*h^{\tiny \text{enc}}_2, \dots, \*h^{\tiny \text{enc}}_L \in \mathbb{R}^D$ is the sequence of outputs of a deep neural network model $\*h^{\tiny \text{enc}}_{\phi}(\*x_{1:T})$ processing the input sequence of $T$ words.
The model $\*h^{\tiny \text{enc}}_{\phi}(\*x_{1:T})$ must then map a sequence of $T$ vectors to a sequence of $L$ vectors.
We define $\*h^{\tiny \text{enc}}_{\phi}$ by a \textit{pretrained} BERT model \citep{devlin2018bert}, followed by a single Transformer block, randomly initialized.
The Transformer block processes the $T$ ($D$-dimensional) outputs from BERT as keys and values, together with a set of $L$ learnable vectors $\*q_{1:L}$ as queries. 
The right hand side of Figure~\ref{fig:model}  illustrates the complete encoder architecture.

\section{On the HSN Training Objective}\label{app:training_objective}

Following standard methods \cite{bishop2006pattern} one readily can show that the Evidence Lower Bound (ELBO) of the Hidden Schema Network model is given by
\begin{multline}    
    \mathcal{L}[\theta, \phi] =  \, \frac{1}{N} \, \sum_{n=1}^N   \Big\{ \\ \mathbb{E}_{q_{\phi}(\*z^{\phantom{(n)}}_{1:L}|\*x^{(n)}_{1:T}, \, \*A) q_{\phi}(\*A)} \Big[\log p_{\theta} (\*x^{(n)}_{1:T} | \*z^{\phantom{(n)}}_{1:L})\Big] \\
     - \mathbb{E}_{q_{\phi}(\*A)} \text{KL}\Big[q_{\phi}(\*z_{1:L}| \*x^{(n)}_{1:T}, \*A); p(\*z_{1:L}| \*A) \Big] \Big\} \\ - \text{KL}[q_{\phi}(\*A);p(\*A)], 
    \label{eq:ELBO-supmat}
\end{multline}
where $\text{KL}[\cdot]$ denotes the Kullback-Leibler (KL) divergence. 

Note that this is \textit{not} the training objective of the main text. 
There we maximize the ELBO together with the mutual information between sentences and schemata. 
We give details about this modified objective in subsections \ref{sec:mutual-info} and \ref{subsec:final_objective} below.
Before getting into that, let us first calculate the explicit expressions for the two divergences above.

\subsection{Kullback-Leibler Divergence Between Random Walks}

For notational convenience we will not write the explicit dependence on the graph $\*A$ in what follows.
Using the explicit product form of the probabilities over walks leads to 

\begin{multline}    
    \text{KL}[q_{\phi}(\*z_{1:T}| \*x^{(n)}_{1:T}); p(\*z_{1:T}) ] =  \\ \sum_{i=2}^L \mathbb{E}_{\hat{q}_{\phi}(\*z_{i-1}| \*x^{(n)}_{1:T}) q_{\phi}(\*z_{i}|\*z_{i-1} \*x^{(n)}_{1:T})} 
     \\ \Big\{ \log \frac{q_{\phi}(\*z_i |\*z_{i-1}, \*x^{(n)}_{1:T})}{ p(\*z_i |\*z_{i-1})} \Big\} + \text{KL}[q_{\phi}(\*z_1); p(\*z_1)], 
    \label{eq:kl-rws}
\end{multline}
where $\hat{q}_{\phi}(\*z_i| \*x^{(n)}_{1:T})$ is the aggregated probability over all walks until step $i$. 
Since the random walks are Markovian, $\hat{q}$ can be explicitly written as
\begin{multline}
    \hat{q}_{\phi}(\*z_i| \*x^{(n)}_{1:T}) = \\ \prod_{1 \le j < i} \*Q^{[j]}(\*x^{(n)}_{1:T}, \phi) \cdot \+\rho( \*x^{(n)}_{1:T}, \phi),
    \label{eq:marginal_walk}
\end{multline}
where the (posterior) transition matrices $\*Q^{[i]}$ are defined in Eq.~\ref{eq:transition_prob_matrix_q} of the main text.
Using the generic random walk definition in Eqs.~\ref{eq:prior_rw} we can readily write the argument of the expectation value in Eq.~\ref{eq:kl-rws} above as
\begin{multline}
    \log \frac{q_{\phi}(\*z_i |\*z_{i-1}, \*x^{(n)}_{1:T})}{ p(\*z_i |\*z_{i-1})} = \\ \sum_{k, j} z^k_{i} z^j_{i-1} \log \frac{Q^{[i-1]}_{k, j}( \*x^{(n)}_{1:T}, \phi)}{P_{k, j}},
\end{multline}
which means we only need to compute the expectation of the product $z^k_{i} z^j_{i-1}$. This one can easily be shown to be
\begin{multline}    
\mathbb{E}_{\hat{q}_{\phi}(\*z_{i-1}| \*x^{(n)}_{1:T}) q_{\phi}(\*z_{i}|\*z_{i-1} \*x^{(n)}_{1:T})} \Big[z^k_{i} z^j_{i-1}\Big] = \\ Q^{[i-1]}_{k,j}( \*x^{(n)}_{1:T}, \phi) \, \hat{\rho}^{[i-1]}_{j}( \*x^{(n)}_{1:T}, \phi),
\end{multline}
where $\hat{\rho}^{[i]}_{j}( \*x^{(n)}_{1:T}, \phi)$ is the $j$th class probability of $\hat{q}_{\phi}(\*z_{i}| \*x^{(n)}_{1:T})$, defined in Eq.~\ref{eq:marginal_walk}.

 Finally, the second KL term in Eq.~\ref{eq:kl-rws} can be directly evaluated
\begin{multline}
   \text{KL}[q_{\phi}(\*z_1); p(\*z_1)] =  \\ \sum_{j=1}^K \rho_{j}( \*x^{(n)}_{1:T}, \phi) \log \frac{\rho_{j}( \*x^{(n)}_{1:T}, \phi)}{\rho_{j}},
\end{multline}
where $\rho_{j}( \*x^{(n)}_{1:T}, \phi)$ and $\rho_{j}$ are, respectively, the posterior and prior class probabilities for the random walks' starting points.

Putting all together we write
\begin{multline}    
    \text{KL}[q_{\phi}(\*z_{1:T}| \*x^{(n)}_{1:T}); p(\*z_{1:T}) ] =  \\ \,  \sum_{i=2}^L \sum_{k, j=1}^K  Q^{[i-1]}_{k,j}( \*x^{(n)}_{1:T}, \phi) \, \hat{\rho}_{j}^{[i-1]}( \*x^{(n)}_{1:T}, \phi) \\ \times \log \frac{Q^{[i-1]}_{k, j}( \*x^{(n)}_{1:T}, \phi)}{P_{k, j}} 
      \\ + \sum_{j=1}^K \rho_{j}( \*x^{(n)}_{1:T}, \phi) \log \frac{\rho_{j}( \*x^{(n)}_{1:T}, \phi)}{\rho_{j}}
\end{multline}

\subsection{Kullback-Leibler Divergence Between Random Graph Models}

Since both prior and posterior graph models treat each edge in ${\cal G}$ as a Bernoulli random variable, we can write directly
\begin{multline}
    \text{KL}[q(\*A); p(\*A)] = \\ \sum_{ij} \left\{p_{\phi}(\*e_i, \*e_j) \log\left(\frac{p_{\phi}(\*e_i, \*e_j)}{p}\right)  \right. \\ \left. + (1-p_{\phi}(\*e_i, \*e_j)) \log \left(\frac{1-p_{\phi}(\*e_i, \*e_j)}{1-p} \right) \right\},
\end{multline}
where $p_{\phi}(\*e_i, \*e_j)$ is the posterior link probability, which is conditioned on the symbols connected by the link, and $p$ is the global prior probability over all links.

\subsection{Maximizing Mutual Information} 
\label{sec:mutual-info}

We would like to maximize the mutual information between the word sequences in our dataset and the schema representations.
We have argued that the training objective in the main text already includes such a mutual information term.
To see this is indeed the case we need to workout some identities.

Let us, for simplicity of notation, consider two discrete variables $\*z$ and $\*x$, the last of which follows an unknown distribution $p_{\cal D}(\*x)$. What follow are identities

\begin{align}
- & \mathbb{E}_{p_{\mathcal{D}}(\*x)}\text{KL}[q(\*z|\*x); p(\*z)]  \nonumber \\
= & \mathbb{E}_{p_{\mathcal{D}}(\*x)}\mathbb{E}_{q(\*z|\*x)} \Big\{ \log p(\*z) - \log (\*z|\*x) \Big\} \nonumber \\ 
= & H_q(\*z|\*x) + \sum_{\*x} \, p_{\mathcal{D}}(\*x) \sum_{\*z} \, q(\*z|\*x) \Big\{ \log p(\*z) \nonumber \\ 
& + \log q^*(\*z) - \log q^*(\*z) \Big\} \nonumber \\
= & H_q(\*z|\*x) - H_{q^*}(\*z) 
\nonumber \\
& + \sum_{\*z} \, q^*(\*z) \Big\{ \log p(\*z) - \log q^*(\*z) \Big\} \nonumber \\
= &-I(\*z; \*x) - \text{KL}[q^*(\*z); p(\*z)],
\label{eq:mutual-info-identities}
\end{align}

where 

\begin{multline}    
    H_q(\*z|\*x) = \\ - \sum_{\*x} \, p_{\mathcal{D}}(\*x) \sum_{\*z} \, q(\*z|\*x) \log q(\*z|\*x),
\end{multline}
is the conditional entropy with respect to distribution $q$ (see e.g. page 17 in \citet{cover1991information}) and

\begin{equation}
    H_{q^*}(\*z) = - \sum_{\*z} \, q^*(\*z) \log q^*(\*z),
\end{equation}
is the entropy of distribution $q^*(\*z)$, which we define as the marginal (data-aggregated) distribution

\begin{equation}
    q^*(\*z) = \sum_{\*x} \, p_{\mathcal{D}}(\*x) q(\*z|\*x).
\end{equation}

Finally, we used the definition of mutual information

\begin{equation}
    I(\*x; \*z) = H_{q^*}(\*z) - H_q(\*z|\*x).  
\end{equation}
See e.g. page 20 in \citet{cover1991information}.

It follows from Eq.~\ref{eq:mutual-info-identities} that maximizing the ELBO (Eq.~\ref{eq:ELBO-supmat}), together with the mutual information between word sequences and schemata, simply amounts to replacing the KL between the approximate posterior and prior random walk distributions, with the KL between the \textit{aggregated posterior} and prior random walk distributions. To wit
\begin{multline} 
    I(\*z_{1:L}; \*x_{1:T}| \*A) -  \frac{1}{N}  \sum_{n=1}^N  \mathbb{E}_{q_{\phi}(\*A)} \Big\{\\KL\Big[q_{\phi}(\*z_{1:L}| \*x^{(n)}_{1:T}, \*A); p(\*z_{1:T}| \*A) \Big] \Big\} \\ =  - \mathbb{E}_{q_{\phi}(\*A)} KL\Big[q^*_{\phi}(\*z_{1:L}|\*A); p(\*z_{1:T}|\*A) \Big], 
    \label{eq: mutual-info}
\end{multline}
where we introduced the aggregated posterior over random walks wrt the word sequence
\begin{multline}
    q^*_{\phi}(\*z_{1:L}|\*A) = \mathbb{E}_{p(\*x_{1:T})} \Big[ q_{\phi}(\*z_{1:L}| \*x_{1:T}, \*A) \Big] \\ \approx \frac{1}{N} \, \sum_{n=1}^N q_{\phi}(\*z_{1:L}| \*x^{(n)}_{1:T}, \*A).
\end{multline}

\begin{table*}[th!]
\centering
\scriptsize
\begin{tabular}{cccccccc} 
\hline
& Dataset  & n. edges  & $\cal D$  & $l$  & $\cal C$  & $\cal CC$  & largest $\cal CC$ \\
\hline 

\hline
 \multirow{4}{*}{HSN(50, 5)} & \texttt{PTB}  &  694.26 $\pm$  9.47  &  2.00 $\pm$  0.00  &  1.43 $\pm$  0.01  &  0.83 $\pm$  0.01  &  1.00 $\pm$  0.00  &  50.00 $\pm$  0.00 \\
 &\texttt{YAHOO}  &  892.67 $\pm$  8.22  &  2.00 $\pm$  0.00  &  1.24 $\pm$  0.01  &  0.84 $\pm$  0.00  &  2.00 $\pm$  0.04  &  49.00 $\pm$  0.04 \\
 &\texttt{YELP}  &  891.06 $\pm$  6.50  &  2.73 $\pm$  0.46  &  1.24 $\pm$  0.03  &  0.84 $\pm$  0.01  &  2.24 $\pm$  0.85  &  48.76 $\pm$  0.85 \\
 &\texttt{Random} 
 &  611.69 $\pm$  17.61 
 &  2.00 $\pm$  0.00 
 &  1.50 $\pm$  0.01 
 &  0.50 $\pm$  0.02 
 &  1.00 $\pm$  0.00 
 &  50.00 $\pm$  0.00 \\
\hline
\multirow{4}{*}{HSN(50, 20)} & \texttt{PTB}  &  764.28 $\pm$  7.88  &  2.83 $\pm$  0.38  &  1.27 $\pm$  0.03  &  0.82 $\pm$  0.01  &  4.71 $\pm$  0.77  &  46.29 $\pm$  0.77 \\
 &\texttt{YAHOO}  &  356.35 $\pm$  7.76  &  3.17 $\pm$  0.37  &  1.57 $\pm$  0.05  &  0.58 $\pm$  0.02  &  12.04 $\pm$  1.49  &  38.96 $\pm$  1.49 \\
 &\texttt{YELP}  &  259.42 $\pm$  5.47  &  2.68 $\pm$  0.48  &  1.42 $\pm$  0.03  &  0.48 $\pm$  0.01  &  20.77 $\pm$  0.68  &  30.23 $\pm$  0.68 \\
 &\texttt{Random} 
 &  611.69 $\pm$  17.61 
 &  2.00 $\pm$  0.00 
 &  1.50 $\pm$  0.01 
 &  0.50 $\pm$  0.02 
 &  1.00 $\pm$  0.00 
 &  50.00 $\pm$  0.00 \\
\hline
\multirow{4}{*}{HLN(100, 5)} & \texttt{PTB}  &  1198.18 $\pm$  16.44  &  2.56 $\pm$  0.50  &  1.76 $\pm$  0.01  &  0.83 $\pm$  0.01  &  1.19 $\pm$  0.40  &  99.81 $\pm$  0.40 \\
& \texttt{YAHOO} &  1239.21 $\pm$  12.19  &  3.15 $\pm$  0.38  &  1.42 $\pm$  0.03  &  0.51 $\pm$  0.01  &  35.93 $\pm$  1.41  &  65.07 $\pm$  1.41 \\
& \texttt{YELP} &  1295.68 $\pm$  12.93  &  3.36 $\pm$  0.48  &  1.55 $\pm$  0.03  &  0.57 $\pm$  0.01  &  27.38 $\pm$  1.69  &  73.62 $\pm$  1.69 \\
& \texttt{Random} & 2474.92 $\pm$  36.58 &  2.00 $\pm$  0.00 &  1.50 $\pm$  0.01 &  0.50 $\pm$  0.01 &  1.00 $\pm$  0.00 &  100.00 $\pm$  0.00 \\
\hline
\multirow{4}{*}{HLN(100, 20)} & \texttt{PTB}  &  892.53 $\pm$  10.04  &  3.04 $\pm$  0.24  &  1.41 $\pm$  0.04  &  0.45 $\pm$  0.01  &  46.04 $\pm$  1.53  &  54.96 $\pm$  1.54 \\
& \texttt{YAHOO} &  261.13 $\pm$  7.14  &  2.18 $\pm$  0.38  &  1.95 $\pm$  0.00  &  0.91 $\pm$  0.01  &  1.01 $\pm$  0.10  &  99.99 $\pm$  0.10 \\
& \texttt{YELP} &  515.84 $\pm$  10.09  &  3.68 $\pm$  0.48  &  1.79 $\pm$  0.06  &  0.38 $\pm$  0.02  &  45.27 $\pm$  2.58  &  55.67 $\pm$  2.58 \\
& \texttt{Random} & 2474.92 $\pm$  36.58 &  2.00 $\pm$  0.00 &  1.50 $\pm$  0.01 &  0.50 $\pm$  0.01 &  1.00 $\pm$  0.00 &  100.00 $\pm$  0.00 \\
\hline
\end{tabular}
\caption{Statistic of inferred graphs for all datasets}
\label{tab:graph_stats_LM_long_sup_mat}
\end{table*} 
In practice we approximate this quantity with
\begin{equation}
    q^*_{\phi}(\*z_{1:L}|\*A) \approx q^*_{\phi}(\*z_1) \prod^{L}_{i=2} \, q^*_{\phi}(\*z_{i}| \*z_{i-1}, \*A),
\end{equation}
where $q^*_{\phi}(\*z_1)$ is a categorical distribution whose class probabilities $\rho^*_j(\phi)$ are the average of those from our approximate posterior, that is
\begin{equation}
    \rho^*_j(\phi) = \frac{1}{N} \, \sum_{n=1}^N \rho_j(\*x_{1:T}^{(n)}, \phi),
\end{equation}
and the transition probabilities $q^*_{\phi}(\*z_{i}| \*z_{i-1}, \*A)$ have transition probability matrices 
\begin{equation}
    Q_{k,j}^{* \, [i]}(\*A, \phi) = \frac{1}{N} \, \sum_{n=1}^N Q_{k,j}^{[i]}(\*x_{1:T}^{(n)}, \*A, \phi).
\end{equation}

\subsection{Training Objective} \label{subsec:final_objective}

Putting everything together, the training objective for the HSN model reads
\begin{multline}    
    \mathcal{L}[\theta, \phi] = \, \frac{1}{N} \, \sum_{n=1}^N \Big\{ \\  \mathbb{E}_{q_{\phi}(\*z^{\phantom{(n)}}_{1:L}|\*x^{(n)}_{1:T}, \, \*A) q_{\phi}(\*A)} \log p_{\theta}(\*x^{(n)}_{1:T} | \*z^{\phantom{(n)}}_{1:L}) \Big\} \\
     - \mathbb{E}_{q_{\phi}(\*A)} \text{KL}\Big[q^*_{\phi}(\*z_{1:L}| \*A); p(\*z_{1:L}| \*A) \Big] \\ - \text{KL}[q_{\phi}(\*A);p(\*A)]. 
    \label{eq:ELBO}
\end{multline}

\subsection{Mean-Field Solution} \label{subsec:MF}

Instead of modeling the posterior over random walks in the usual way, we could consider a mean-field decomposition along the time component, by ignoring the dependency on the graph $\mathcal{G}$
\begin{equation} \label{eq:MF}
    q_{\phi}(\*z_{1:L}| \*x_{1:T}) = \prod_{i=1}^L q_{\phi}(\*z_i | \*x_{1:T}),
\end{equation}
where at each step of the walk we have a step-dependent categorical distribution
\begin{equation}
    q_{\phi}(\*z_i | \*x_{1:T}) = \prod_{j=1}^K \left(\rho^{[i]}_j (\*x_{1:T}, \phi) \right)^{z_i^j},
\end{equation}
whose class probabilities live in the $K$-simplex.
We could model the latter via
\begin{equation}
   \+\rho^{[1]}, \dots, \+\rho^{[L]} = \text{softmax}(\*h_1^{\tiny \text{enc}}, \dots, \*h_L^{\tiny \text{enc}})
\end{equation}
where $\*h_1^{\tiny \text{enc}}, \dots, \*h_L^{\tiny \text{enc}}$ are the outputs of our encoder neural network model, shown in Figure~\ref{fig:model} of the main text.

Replacing the mean-field approximation of Eq.~\ref{eq:MF} into Eq.~\ref{eq:ELBO-supmat} yields 
\begin{align}
    & \text{KL}L[q_{\phi}(\*z_{1:T}| \*x^{(n)}_{1:T}); p(\*z_{1:T}|\*A) \nonumber \\
    =&   \sum_{i=2}^L \Big\{ \mathbb{E}_{q_{\phi}(\*z_{i}| \*x^{(n)}_{1:T})} \log q_{\phi}(\*z_i |\*x^{(n)}_{1:T}) \nonumber \\ 
    & - \mathbb{E}_{q_{\phi}(\*z_{i}| \*x^{(n)}_{1:T}) q_{\phi}(\*z_{i-1}| \*x^{(n)}_{1:T})}\log p(\*z_i |\*z_{i-1}) \Big\} \nonumber \\ 
    & + \text{KL}[q_{\phi}(\*z_1); p(\*z_1)], \nonumber \\
    =&  \sum_{i=1}^L  \sum_j^K \rho^{[i]}_j(\*x_{1:T}, \phi) \log \frac{\rho^{[i]}_j(\*x_{1:T}, \phi)}{\rho_j}  \nonumber \\ 
    & - \sum_{i=2}^L \sum_{k, j}^K \mathbb{E}_{q_{\phi}(\*z_{i}| \*x^{(n)}_{1:T}) q_{\phi}(\*z_{i-1}| \*x^{(n)}_{1:T})} \Big[ z_i^k z_{i-1}^j \Big] \nonumber \\
    & \times \log P_{k, j} \nonumber \\
    =&  \sum_{i=1}^L  \sum_j^K \rho^{[i]}_j(\*x_{1:T}, \phi) \log \frac{\rho^{[i]}_j(\*x_{1:T}, \phi)}{\rho_j} \nonumber\\ 
    &- \sum_{i=2}^L \sum_{k, j}^K \rho^{[i]}_k(\*x_{1:T}, \phi) \rho^{[i-1]}_j(\*x_{1:T}, \phi) \nonumber \\
    & \times \log P_{k, j}. 
    \label{eq:kl-rws-mean-field}
\end{align}

\subsection{Fully Connected Graph}

We can replace the adjacency matrix $\*A$ in the definition of the transition probability matrix of our posterior $\*Q(\*x_{1:T}, \*A, \phi)$, with that of a fully connected graph. 
The aggregated posterior over all walks up to step $i$ (Eq.~\ref{eq:marginal_walk} above) reduces in this case to

\begin{multline}    
    \hat{\rho}_k^{[i]} (\*x_{1:T}, \phi)  = \\ \sum_j^K \left( \frac{f_k^{[i-1]}(\*x_{1:T}, \phi) A_{k, j}}{\sum_m f_m^{[i-1]}(\*x_{1:T}, \phi) A_{m, j}}\right) \hat{\rho}_j^{[i-i]} (\*x_{1:T}, \phi) \\
     =   \left( \frac{f_k^{[i-1]}(\*x_{1:T}, \phi)}{\sum_m f_m^{[i-1]}(\*x_{1:T}, \phi)}\right) \left( \sum_j^K \hat{\rho}_j^{[i-i]} (\*x_{1:T}, \phi) \right) \\ =  \frac{f_k^{[i-1]}(\*x_{1:T}, \phi)}{\sum_m f_m^{[i-1]}(\*x_{1:T}, \phi)},
\end{multline}
which is equivalent to that of the mean-field approximation of section \ref{subsec:MF} with $\hat{\rho}_k^{[i]} = \rho_k^{[i]}$.

\section{On Synthetic Dataset Experiments}
\label{app:synthetic}

In this section we give additional details of and results from our proof-of-concept experiments.

\subsection{Synthetic Language Model}
We generate our synthetic dataset as follows:
first, we sample a single, fixed graph ${\cal G^*}$ with $K$ nodes from a predefined random graph model.
Second, we define a set of random tokens ${\cal V}$, of size $V$, to be our vocabulary. We create each token as a random 3-tuple from the Latin alphabet, and choose to have at least one order of magnitude more tokens than nodes in $G$ (that is, $V \gg K$).
Third, we assign a random bag of tokens to each node in ${\cal G^*}$. 
These random bags can simply be understood as probability distributions over ${\cal V}$, and can be represented as $V$-dimensional vectors whose components live on the simplex.
Note in particular that, by construction, tokens can be shared among the different nodes of ${\cal G^*}$.
Finally, let us identify the $K$ random bags with the $K$ symbols $\{\*e_1, \*e_2, \dots, \*e_K\}$ of the synthetic language model.

To generate synthetic sentences we sample uniform, $L$-step random walks on ${\cal G^*}$, whose transition matrix is given by Eq.~4 in the main text, with $\*f = \mathbb{I}$.
Having obtained a set of random walks on ${\cal G^*}$, we sample one random token from each of the symbols (i.e. from each random bag) along the walks.

\subsection{Experimental Settings}

Here we give additional details for reproducibility

\textbf{Datasets}
\begin{itemize}
    \item Following the procedure above we generated two datasets from two random graphs with different topologies. One sampled from the Barab\'asi-Albert model \citep{barabasi1999emergence}, the other from the Erd\"os-R\'enyi model \citep{erdosRenyi}. 
We generate these graphs using NetworkX, a Python language software package for network structures \citep{SciPyProceedings_11}. 
Specifically, we generate Barab\'asi-Albert graphs by attaching 3 edges from each new node to old ones, and Erd\"os-R\'enyi graphs with an edge probability of 0.5.
We set both graphs to have $K=100$ symbols.
 
 \item We define each random bag of tokens in $\cal G^*$ to have two tokens only (each with equal probability). 
 
 \item We use a vocabulary of 1000 random tokens.
 
 \item Once the graph is fixed, we set the token sequence length to $L=10$ ($L=11$) for the Erd\"os (Barab\'asi) datasets
and generate a total of $N=100000$ token sequences from each random graph.
\end{itemize}

 \textbf{Hidden Schema Network} (HSN) \textbf{settings}
 \begin{itemize}
     \item We train randomly initialized embeddings of dimension 256, one for each token. We sample these from a normal distribution with zero mean and a standard deviation of 0.01.
     \item The posterior graph model is defined via a single feed-forward neural network with 256 hidden units.
     \item The prior graph model has the edge probability $p$ as hyperparameter. We crossvalidate it from the set $p=\{ 0.1, 0.2, 0.5, 0.6, 0.8 \}$ and found that HSN could fit the Barab\'asi dataset only with small values $\{0.1, 0.2\}$. HSN could fit the Erd\"os dataset with larger values $\{0.5, 0.6\}$
     \item The posterior random walk model is defined by replacing BERT with a 2-block Transformer encoder \citep{vaswani2017attention}, each with 2 heads, 256 hidden units and dropout probability of 0.2.
     \item The prior random walk model was set to a uniform random walk. 
 \end{itemize}
 
 \textbf{Training details}
 
 \begin{itemize}
     \item We use a batch size of 256 and train with Adam \citep{kingma2014adam}, with a learning rate of 0.0001, in all experiments.
     \item To sample both graph and random walk posterior models with use the Gumbel-Softmax trick \citep{jang2016categorical}, with a constant temperature of 0.75
     \item We train the models for 200 epochs
 \end{itemize}

\subsection{Additional Results}
 
 \begin{table*}[t!]
\centering
\scriptsize
\begin{tabular}{|c |c| c c  c |c c c c|} 
    \hline
Graph ${\cal G}^*$ & Model & NLL & $\text{KL}-{\*z}$ & $\text{KL}-{\cal G}$ & AUC & $|{\cal G}^*-{\cal G}|_F$ & $|{\cal G}^{\tiny \text{r}}-{\cal G}|_F$ & N. edges(${\cal G}$)    \\
    \hline \hline
\multirow{3}{*}{Barabasi} & LSTM   & \textbf{53.07}$\pm$ \textbf{0.01} & -- & -- & -- & -- & -- & -- \\     
&HS ($0.1$)   & 53.08 $\pm$ 0.01 & 0.10 $\pm$ 0.06 & 9 $\pm$ 1   & 0.977 $\pm$ 0.003 & 17 $\pm$ 2& 27 $\pm$ 1& 1090 $\pm$ 143\\     
&HS ($0.2$)   & 53.07 $\pm$ 0.02 & 0.09  $\pm$ 0.06  & 4.8 $\pm$ 0.5 & 0.989 $\pm$ 0.001   & 17 $\pm$ 2 & 26 $\pm$ 1 & 1360 $\pm$ 104 \\   
    \hline \hline
\multirow{3}{*}{Erdos} &LSTM   & \textbf{48.24} $\pm$ \textbf{0.02} & -- & -- & -- & -- & -- & -- \\    
&HS ($0.5$)   & 50.9 $\pm$ 0.8   & 1.2 $\pm$ 0.3   & 4 $\pm$ 6   & 0.95 $\pm$ 0.06   & 34.8 $\pm$ 0.9& 40 $\pm$ 5& 2812 $\pm$ 344\\     
&HS ($0.6$)   & 50.4 $\pm$ 0.6   & 1.3 $\pm$ 0.1   & 1 $\pm$ 2   & 0.94 $\pm$ 0.06   & 36.8 $\pm$ 0.8& 44 $\pm$ 2& 3131 $\pm$ 156\\     
    \hline
\end{tabular}
\caption{Inference on ground-truth random graphs. Here we use the notation HS($p$) to denote Hidden Schema Network models with prior graph distributions whose edge probability is set to $p$.}
\label{tab:results_synthetic_sup_mat}
\end{table*}

 Table~\ref{tab:results_synthetic_sup_mat} displays the mean and standard deviation of some additional results on our proof-of-concept experiments.
We trained ten models in total.
 
 We first trained a simple LSTM Network to infer the correct symbol order in each random token sequence. 
We noticed that a network with 256 hidden units was enough to solve this task perfectly.
Indeed, the negative log-likelihood (NLL) of these models corresponds to choosing the 2-token random bag sequence (i.e. the schema) that yields the correct token sequence without errors. 
The HSN performs equally well on the Barab\'asi dataset, and slightly worst on the Erd\"os dataset.
In fact, we have noticed the Erd\"os dataset proved to be more challenging to learn with the HSN in all regards. See, for example, the AUC scores or the Frobenious norms of HSN in this dataset, as compared to the Barab\'asi case.
We think this might be due to the fact that Barab\'asi graphs have more structure, simply because of their sparsity, which arguably make them easier to infer with our inductive bias.

 Note also how increasing the prior edge probability $p$ affects the average number of edges of the inferred graphs.

 \section{On Language Modelling Experiments}
 \label{app:additional_results}
 
 In this section we give additional details of and results from our language modelling and representation learning experiments.
 
 \subsection{Experimental Settings}
 
 Here we give additional details for reproducibility
 
 \textbf{Datasets}
 
 \begin{itemize}
     \item We consider three widely used public datasets, namely the Penn Treebank (PTB) 
\citep{marcus-etal-1993-building}, Yahoo and Yelp \citep{DilatedCNNVAE} corpora.

\item PTB training set has a total of 38219 sentences. The average length of which is of about 22 words. The validation and test set have 5527 and 5462 sentences, respectively.  The minimum (maximum) sentence length in PTB is of 2 (78) words. 
\item Yahoo training set has a total of 100000 sentences. The average length of which is of about 80 words. The validation and test sets have 10000 sentences each.  The minimum (maximum) sentence length in Yahoo is of 21 (201) words. The Vocabulary size is of 200000 words. 
\item Yelp training set has a total of 100000 sentences. The average length of which is of about 97 words. The validation and test sets have 10000 sentences each.  The minimum (maximum) sentence length in Yelp is of 21 (201) words. The Vocabulary size is of 90000 words. 
 \end{itemize}

\textbf{HSN settings}
\begin{itemize}
    \item In all experiments we leveraged pretrained BERT and GPT-2 models, both with 12 layers, 768 hidden dimensions ($D$) and 12 attention heads. We used the public HuggingFace implementation of both these models \citep{wolf-etal-2020-transformers}.
    \item The posterior graph model is set to a 2-layer feed forward network, each with hidden dimension 512.
    \item We crossvalidated the prior edge probability over the set of values $p=\{0.1, 0.2, 0.5, 0.6\}$ and found $p=0.5$ (a maximum entropy prior) to yield the best results. All results we report correspond to this ($p=0.5$) case.
    \item We also train an inhomogeneous random walk prior model by making $\+\rho$ and the sequence of weights $\*f^{[1]}, \*f^{[2]}, \dots, \*f^{[L-1]}$ trainable. We initialized them by sampling from a normal distribution with zero mean and standard deviation of 0.01.
    \item We experimented with HSN of $K=\{50, 100\}$ symbols and random walks of length $L=\{5, 20\}$.
\end{itemize}

\textbf{Training details}
\begin{itemize}
    \item We used a batch size of 32 and train with Adam \citep{kingma2014adam}, with a learning rate of 0.00001, in all experiments.
     \item To sample both graph and random walk posterior models with used the Gumbel-Softmax trick \citep{jang2016categorical}, with a constant temperature of 1.0.
     \item We used a cyclical schedule to anneal both KL terms in our training objective from zero to one \citep{fu-etal-2019-cyclical}. 
When the annealing weight (usually called $\beta$ in the literature) is finite, we used a KL threshold scheme \citep{li2019surprisingly}, with a threshold value of 0.1.
     \item We trained the models for 100 epochs, although they usually needed about 60 epochs only to converge (in the NLL).
     \item We applied word dropout to the input of the decoder model with probability 0.3 in the following cases: (i) for all models trained on PTB; (ii) and all models with $L=50$ trained on all datasets.
\end{itemize}

\subsection{Evaluation Metrics}

\subsubsection*{Monte Carlo Perplexity Estimation}
We compute the perplexity of HSN models via the importance weighted Monte Carlo estimator of the marginal log-likelihood,
\begin{multline}
    \log p_\theta(\*x_{1:T}) = \log \frac{1}{R} \sum^R_r  \frac{1}{S} \sum^S_s \\
    \frac{p_\theta(\*x_{1:T} | \*z^{(r)}_{1:L}) p(\*z^{(r)}_{1:L} | \*A^{(s)}) p(\*A^{(s)})}{q_\phi(\*z^{(r)}_{1:L} | \*A^{(s)})q_\phi(\*A^{(s)})},
\end{multline}
where $R$ and $S$ are the number of random walk samples and graph samples, respectively.
We used $R = 100$ and $S = 10$ per each sentence in the test set to estimate the perplexity.

\subsubsection*{Mutual Information}

Another metric that we use to evaluate our models is the mutual information between the representations (random walks) and the word sequences

\begin{multline}
    I(\*z; \*x) = \\ \mathbb{E}_{p_{\mathcal{D}}(\*x)}\text{KL}[q(\*z|\*x); p(\*z)] - \text{KL}[q^*(\*z); p(\*z)].
\end{multline}

Full derivation of this expression can be found in Appendix \ref{sec:mutual-info} above.

 \subsection{Additional Results}\label{subsec:app_addional_results}

\begin{table*}[th!]
\centering
\begin{tabular}{ ll  cccc} 
\hline
 & \footnotesize Model  & \footnotesize MI & \footnotesize Rec & \footnotesize KL & \footnotesize $\text{KL}_{\cal G}$ \\
\hline
& \footnotesize $\text{iVAE}_{\tiny \text{MI}}$  & 
\footnotesize 12.50 & \footnotesize 74.69 & \footnotesize  12.51 & -- \\
& \footnotesize $\text{Optimus}_{ \tiny \text{A}}$ &         
\footnotesize 3.78 & \footnotesize 86.43 & \footnotesize  4.88 & --  \\
& \footnotesize $\text{Optimus}_{ \tiny \text{B}}$  & 
\footnotesize 8.18 & \footnotesize 77.65 & \footnotesize  28.50 & -- \\ 
\texttt{\footnotesize PTB} & \footnotesize $\text{HSN}_{ \tiny \text{50,20}}$  & 
\footnotesize 25.3 $\pm$ 0.9 & \footnotesize 77.1  $\pm$ 0.4 & \footnotesize  27 $\pm$ 1 & \footnotesize 0.1 $\pm$ 0.2 \\
& \footnotesize $\text{HSN}_{ \tiny \text{50,5}}$  & 
\footnotesize 9.3 $\pm$ 0.2 & \footnotesize 78.7 $\pm$ 0.2 & \footnotesize  10.3 $\pm$ 0.2 & \footnotesize 0.02 $\pm$ 0.01 \\
& \footnotesize $\text{HSN}_{ \tiny \text{100,20}}$  & 
\footnotesize 23 $\pm$ 1 & \footnotesize 77.9 $\pm$ 0.3 & \footnotesize  24 $\pm$ 1 & \footnotesize 0.21 $\pm$ 0.05 \\
& \footnotesize $\text{HSN}_{ \tiny \text{100,5}}$  & 
\footnotesize 9.1 $\pm$ 0.3 & \footnotesize 79.1 $\pm$ 0.2 & \footnotesize  10.2 $\pm$ 0.3 & \footnotesize 0.11 $\pm$ 0.02 \\
\hline
& \footnotesize $\text{iVAE}_{\tiny \text{MI}}$  & 
\footnotesize 10.70 & \footnotesize 297.70 & \footnotesize  11.40 & -- \\
& \footnotesize $\text{Optimus}_{ \tiny \text{A}}$ &     
\footnotesize 5.34 & \footnotesize 282.84 & \footnotesize  6.97 & -- \\
& \footnotesize $\text{Optimus}_{ \tiny \text{B}}$  & 
\footnotesize 9.18 & \footnotesize 270.80 & \footnotesize  30.41 & -- \\
\texttt{\footnotesize YAHOO} & \footnotesize $\text{HSN}_{ \tiny \text{50,20}}$  & 
\footnotesize 25 $\pm$ 2 & \footnotesize 267 $\pm$ 1 & \footnotesize  26 $\pm$ 3 & \footnotesize 0.07 $\pm$ 0.1 \\
& \footnotesize $\text{HSN}_{ \tiny \text{50,5}}$  & 
\footnotesize 10.2 $\pm$ 0.4 & \footnotesize 268.4 $\pm$ 0.4 & \footnotesize  11.4 $\pm$ 0.5 &  \footnotesize 0.08 $\pm$ 0.04 \\
& \footnotesize $\text{HSN}_{ \tiny \text{100,20}}$  & 
\footnotesize 13 $\pm$ 5 & \footnotesize 268 $\pm$ 1 & \footnotesize  13 $\pm$ 5 & \footnotesize 0.3 $\pm$ 0.1 \\
& \footnotesize $\text{HSN}_{ \tiny \text{100,5}}$  & 
\footnotesize 10.8 $\pm$ 0.4 & \footnotesize 267.7 $\pm$ 0.5 & \footnotesize  12.3 $\pm$ 0.6 & \footnotesize 0.04 $\pm$ 0.04 \\
\hline
& \footnotesize $\text{iVAE}_{\tiny \text{MI}}$  & 
\footnotesize 11.00 & \footnotesize 348.70 & \footnotesize  11.60 & -- \\
& \footnotesize $\text{Optimus}_{ \tiny \text{A}}$ &     
\footnotesize 2.54 & \footnotesize 334.31 & \footnotesize  3.09 & -- \\
& \footnotesize $\text{Optimus}_{ \tiny \text{B}}$  & 
\footnotesize 9.13 & \footnotesize 325.77 & \footnotesize  27.89 & -- \\
\texttt{\footnotesize YELP} & \footnotesize $\text{HSN}_{ \tiny \text{50,20}}$  & 
\footnotesize 24 $\pm$ 2 & \footnotesize 312 $\pm$ 3 & \footnotesize  26 $\pm$ 3 &  \footnotesize 0.1 $\pm$ 0.1 \\
& \footnotesize $\text{HSN}_{ \tiny \text{50,5}}$  & 
\footnotesize 10.0 $\pm$ 0.2 & \footnotesize 312.0 $\pm$ 0.5 & \footnotesize  11.3 $\pm$ 0.2 &  \footnotesize 0.025 $\pm$ 0.009 \\
& \footnotesize $\text{HSN}_{ \tiny \text{100,20}}$  & 
\footnotesize 18 $\pm$ 3 & \footnotesize 312.7 $\pm$ 0.6 & \footnotesize  17 $\pm$ 3 & \footnotesize 0.3 $\pm$ 0.1 \\
& \footnotesize $\text{HSN}_{ \tiny \text{100,5}}$  & 
\footnotesize 10.1 $\pm$ 0.2 & \footnotesize 312.1 $\pm$ 0.5 & \footnotesize  11.5 $\pm$ 0.3 & \footnotesize 0.16 $\pm$ 0.02 \\
\hline
\end{tabular}
\caption{Mutual information (MI), reconstruction loss (Rec), KL of the random walks (KL) and KL of the graph ($\text{KL}_{\cal G}$) for iVAE.
$\text{Optimus}_{ \tiny \text{A, B}}$ label models with best PPL and MI, respectively (with $\lambda=0.05, 1$) \citep{li-etal-2020-optimus} .
$\text{iVAE}_{\tiny \text{MI}}$ was taken from \citet{fang-etal-2019-implicit}. For Hidden Schema Network models $\text{HSN}_{\text{K,L}}$ with K symbols and random walk length L, we report mean and standard deviation obtained by repeating each experiment five times with different initializations.
}
\label{tab:results_LM_long_sup_mat}
\end{table*}

 Here we report results complementing the conclusions of the main text.
 
 \textbf{Language modelling}. Table~\ref{tab:results_LM_long_sup_mat} displays mutual information, reconstruction loss, KL of the random walks and KL of the graph. We report their mean and standard deviation obtained when repeating the experiments with the HSN model five times, with different initializations.
 
 The conclusion of the main text, viz. that our results outperform all baselines, remains unaltered, even within error bars.
 
 \textbf{Graph statistics}. We characterize the structure of ${\cal G}$ in terms of five statistics: 
(i) the diameter ${\cal D}$, which measures the maximum path length over all node pairs in ${\cal G}$;
(ii) the average distance $l$, which instead measures the average shortest path length between all node pairs;
(iii) the clustering coefficient ${\cal C}$, which represents the probability that two neighbors of a randomly chosen node are themselves neighbors;
(iv) the number of connected components ${\cal CC}$;
and (v) the degree distribution $P(k)$, which represents the probability that a randomly chosen node will have $k$ neighbors.

 Table~\ref{tab:graph_stats_LM_long_sup_mat} reports the statistics of our inferred graphs for all datasets, and all model configurations. 
 
 We can see that increasing the random walk length from 5 to 20 increases the number of connected components of the graphs. As a consequence, subsets of word sequences are map onto smaller subgraphs, the larger of which is about 50 symbols.
One could argue that, since longer random walk lengths imply a larger set of possible schema configurations, the number of symbols required to describe our three corpora can simply decrease.
In other words, less symbols are needed by long schemata.
Similarly, directly increasing the symbols number leads too to a larger number of connected components. Indeed, even the short schemata in Yelp and Yahoo do not use all available symbols to model the corpora. 

\begin{figure*}
\begin{tikzpicture}
\node (a) at (0,0){
        \includegraphics[width=0.45\textwidth]{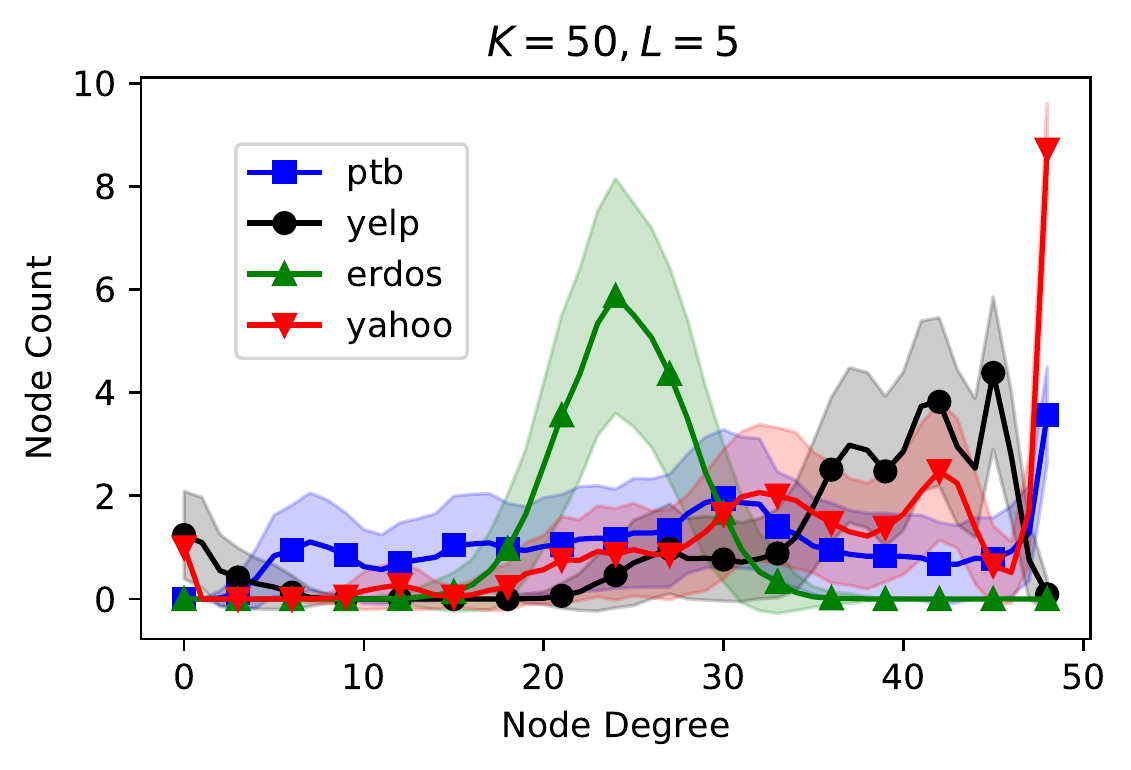}
        }; 
        
\node[anchor=west] (b) at (a.east) {
        \includegraphics[width=0.45\textwidth]{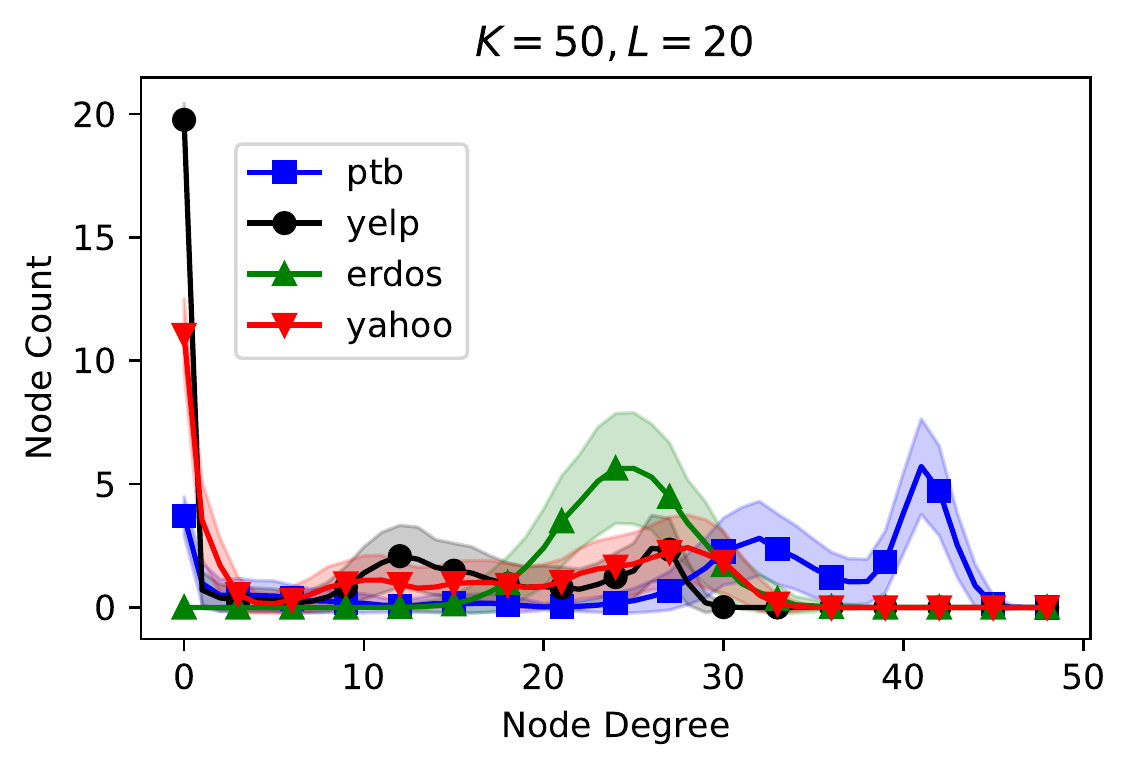}};
    
\node[anchor=north] (c) at (a.south) {
        \includegraphics[width=0.45\textwidth]{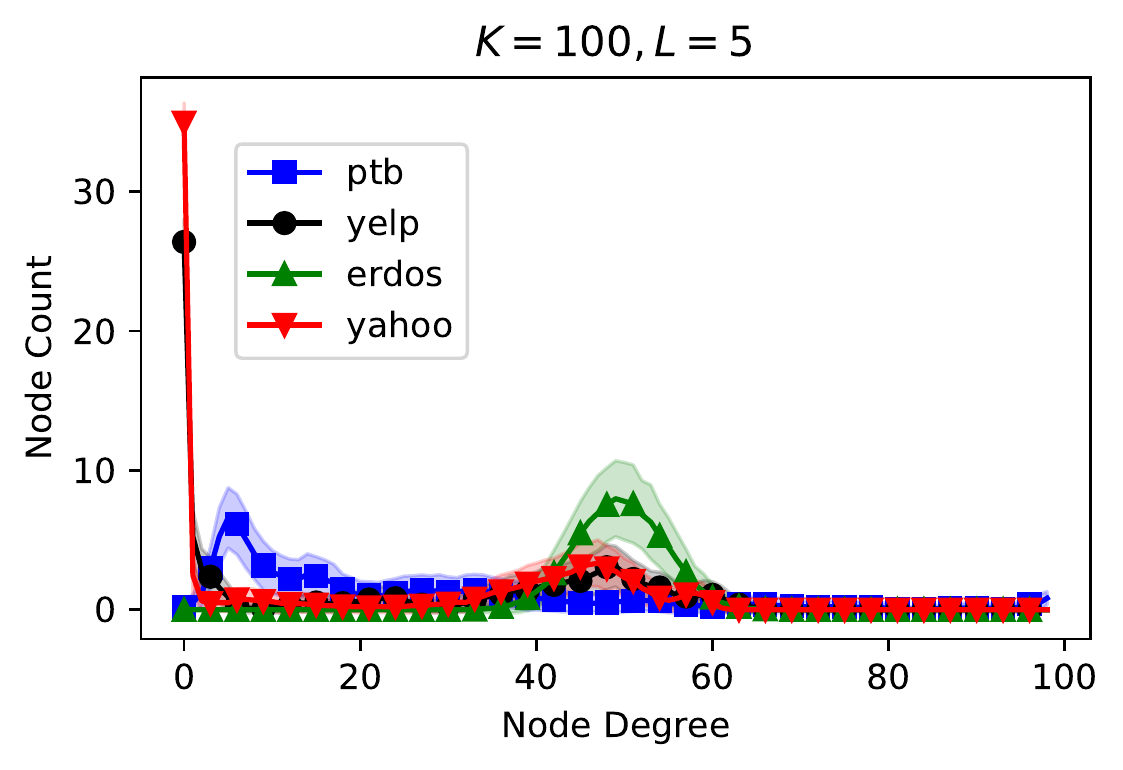}
        };
        
\node[anchor=north] (d) at (b.south) {    
        \includegraphics[width=0.45\textwidth]{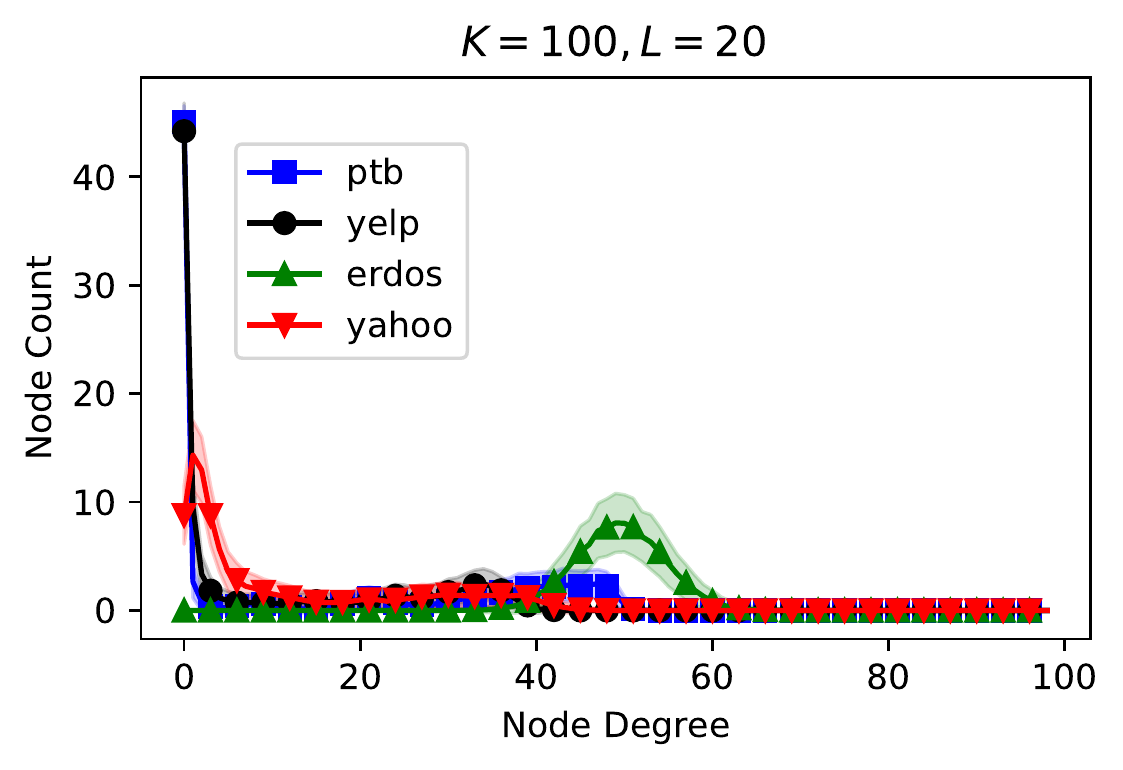}
        };
        
    \node[anchor=north] (e) at (c.south) {    
\includegraphics[width=0.45\textwidth]{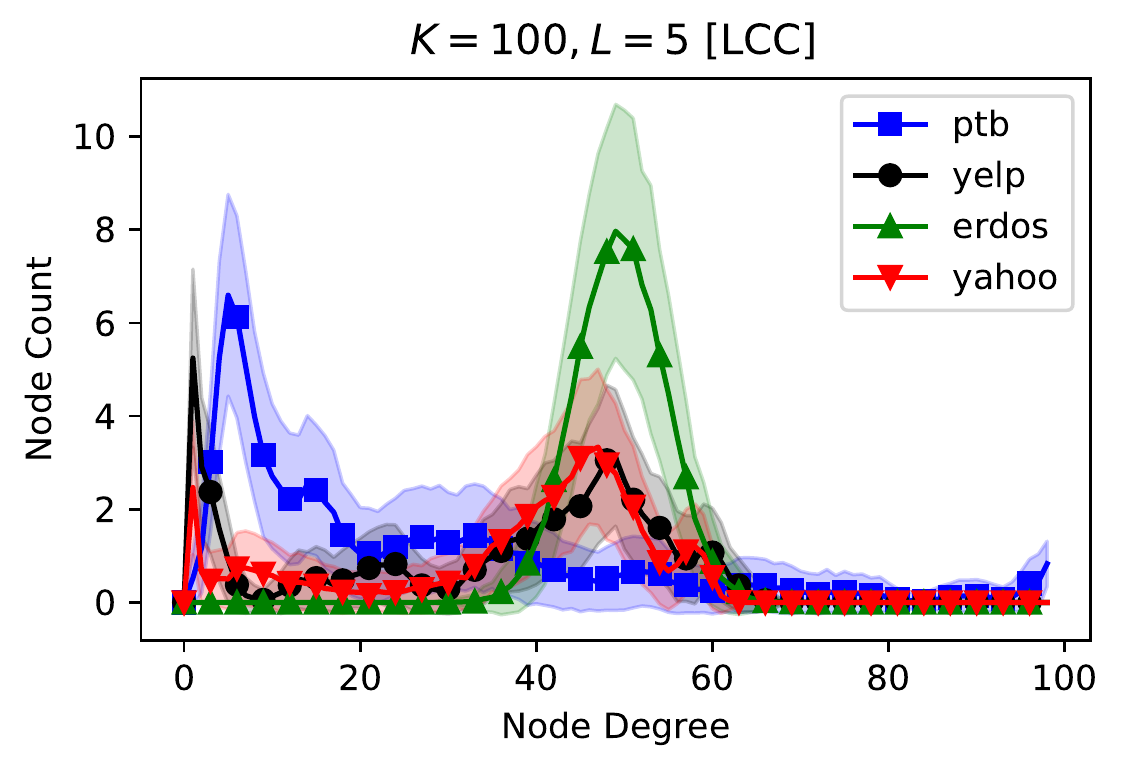}
        };
    
    \node[anchor=north] (f) at (d.south) {     
\includegraphics[width=0.45\textwidth]{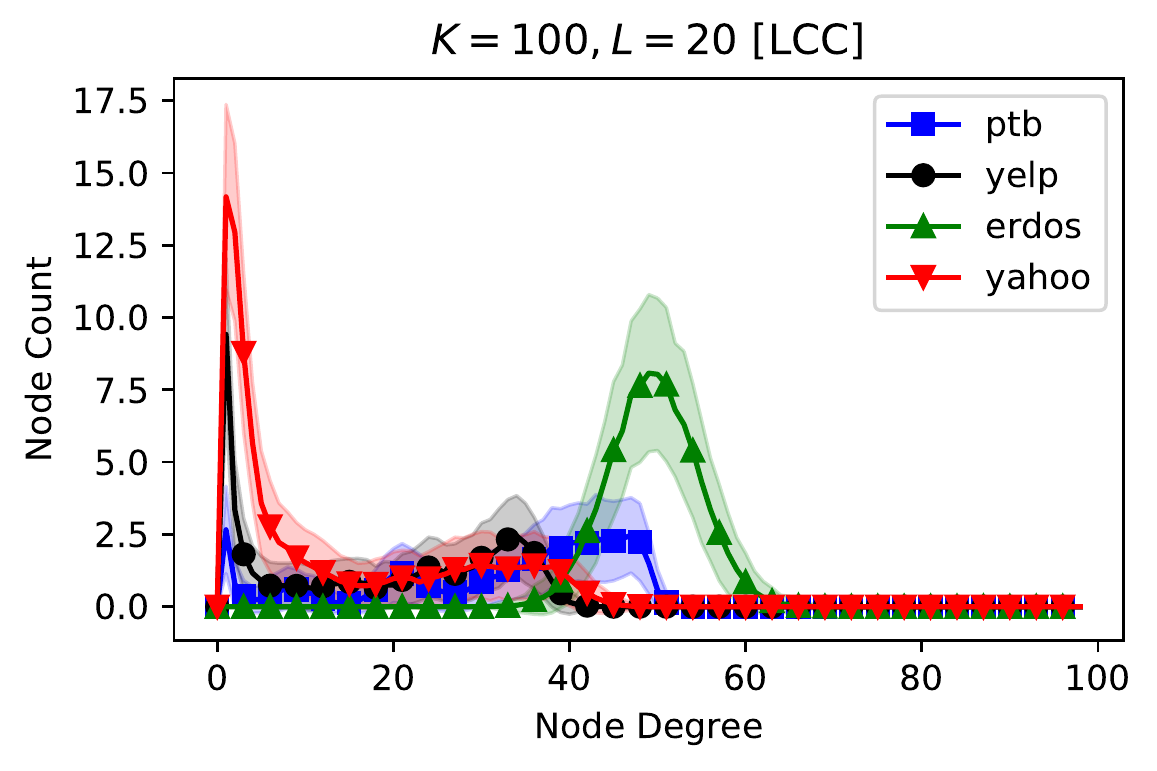}
        };
        
\end{tikzpicture}

\caption{Degree distributions of inferred graphs from all corpora, compared to Erdos-Renyi graphs for $p=0.5$. 
The upper four plots show results for full inferred graphs, the lowest two show the degree distributions of the largest connected component of the models for $K=100$.}\label{fig:degreedist}

\end{figure*}

Finally, Figure \ref{fig:degreedist}  displays the difference between the degree distributions of the schema networks, and that of an Erd\"os-R\'enyi graph for all HSN configurations.

 \textbf{Schema distributions on inferred networks}. We can get a graphical picture of the features we just discussed above in Figures~\ref{fig:yahoo-50}--\ref{fig:yelp} below.
Importantly, we see that the schema distribution is different for each category of each corpora in all model configurations. In other words, we do not observe any kind of mode collapse.

 \textbf{Schema interpolations}. Finally, we have also explored ``schema interpolations'': given two schemata $e_{j_1:j_L}$ and $e_{m_1:m_L}$, we find the shortest path (of length $l$) on ${\cal G}$ connecting the end of $e_{j_1:j_L}$ with the beginning of $e_{m_1:m_L}$. 
Our interpolation steps are the schemata $\{e_{j_1+i:j_L+i} : \forall \, \, 0 \le i \le l+L \, \, \text{along the path}\}$.
Tables~\ref{tab:interpol-supmat-2}--\ref{tab:interpol-supmat-3} show interpolations of random instances from all datasets. Note how the model successfully interpolates between categories in both Yelp and Yahoo.

\section{Which Symbols do Words Attend to? A Preliminary Study on Yelp Reviews}\label{sec:attention_to_symbols}

\begin{table}[th]
    \footnotesize
    \centering
    \begin{tabular}
    {lm{0.21\textwidth}m{0.25\textwidth}m{0.21\textwidth}}
        layer & KL\newline (\textit{good}, \textit{bad}) & KL\newline (\textit{good}, \textit{great}) & KL\newline (\textit{great}, \textit{bad})\\
        \hline
        1 & 0.807 & 0.336 & 1.227\\
        5 & 0.738 & 0.177 &  1.245\\
        12 & 0.635 & 0.224 & 0.957\\
    \end{tabular}
    \caption{Kullback-Leibler divergence between the distributions of most attended symbols, when generating the tokens \textit{good}, \textit{bad} and \textit{great}. Results are computed with HSN(100, 5) trained on Yelp. The KL values are computed for each head separately and then averaged.}
    \label{tab:attn_kl}
\end{table}
 
In this section we investigate how symbols are used by HSN when generating text. We do this by exploring the decoder attention matrix between the symbols and the generated tokens. 
Reading the attention weights, we can examine which symbols are most important for the generation of any given token, i.e. which symbols are attended to more strongly.
A bit more in detail we select, for a given token in a given sentence, the symbol with the highest attention value. 
We can then compute the distribution of most attended symbols when generating that token for the complete dataset.

\begin{figure*}
\centering
\begin{subfigure}[b]{\textwidth}
   \includegraphics[width=1\linewidth]{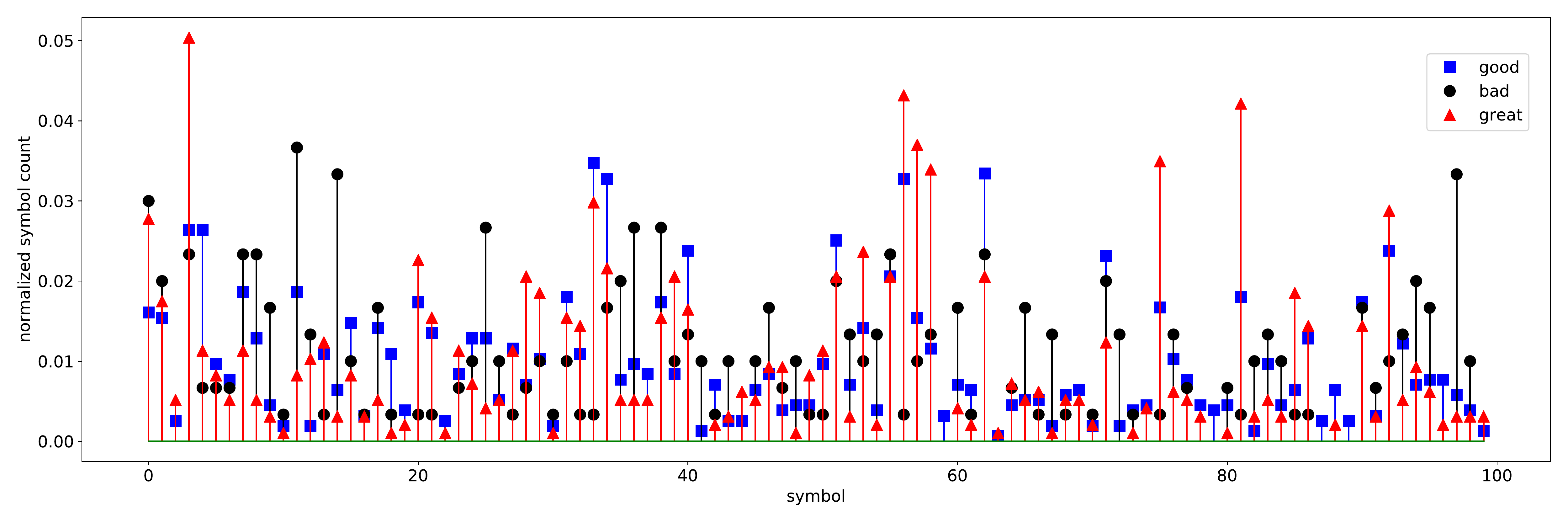}
  \caption{Layer 1}
   \label{fig:attn_avg_layer1} 
\end{subfigure}

\begin{subfigure}[b]{\textwidth}
   \includegraphics[width=1\linewidth]{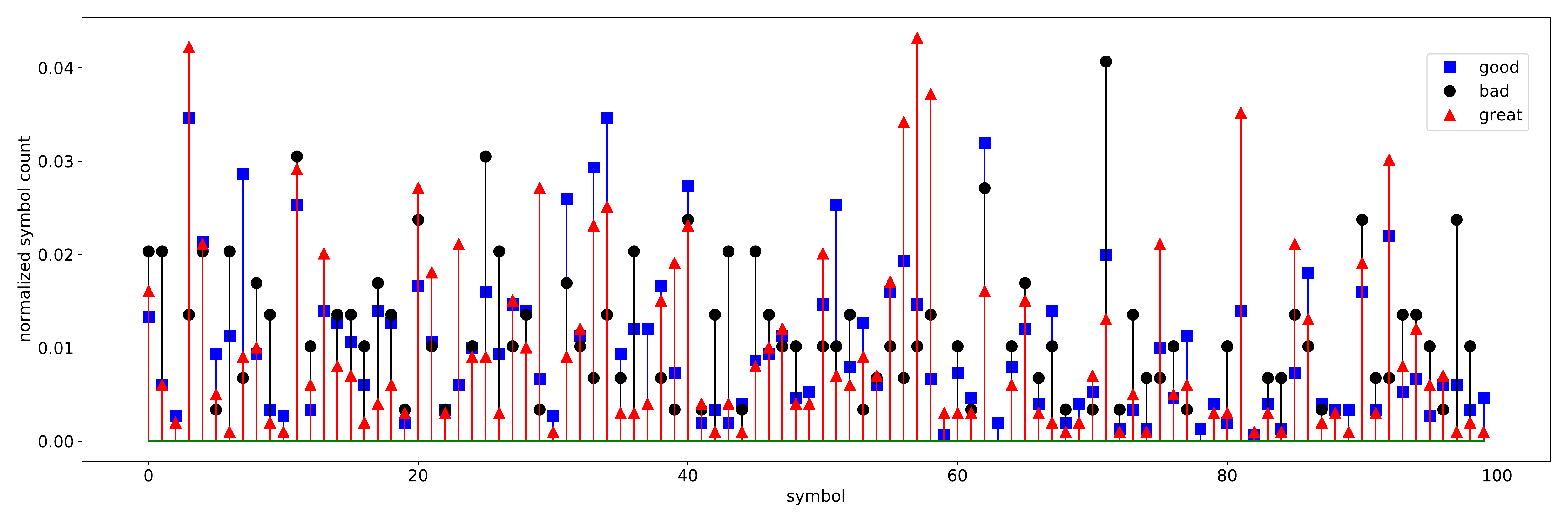}
\caption{Layer 5}
   \label{fig:attn_avg_layer5}
\end{subfigure}

\begin{subfigure}[b]{\textwidth}
   \includegraphics[width=1\linewidth]{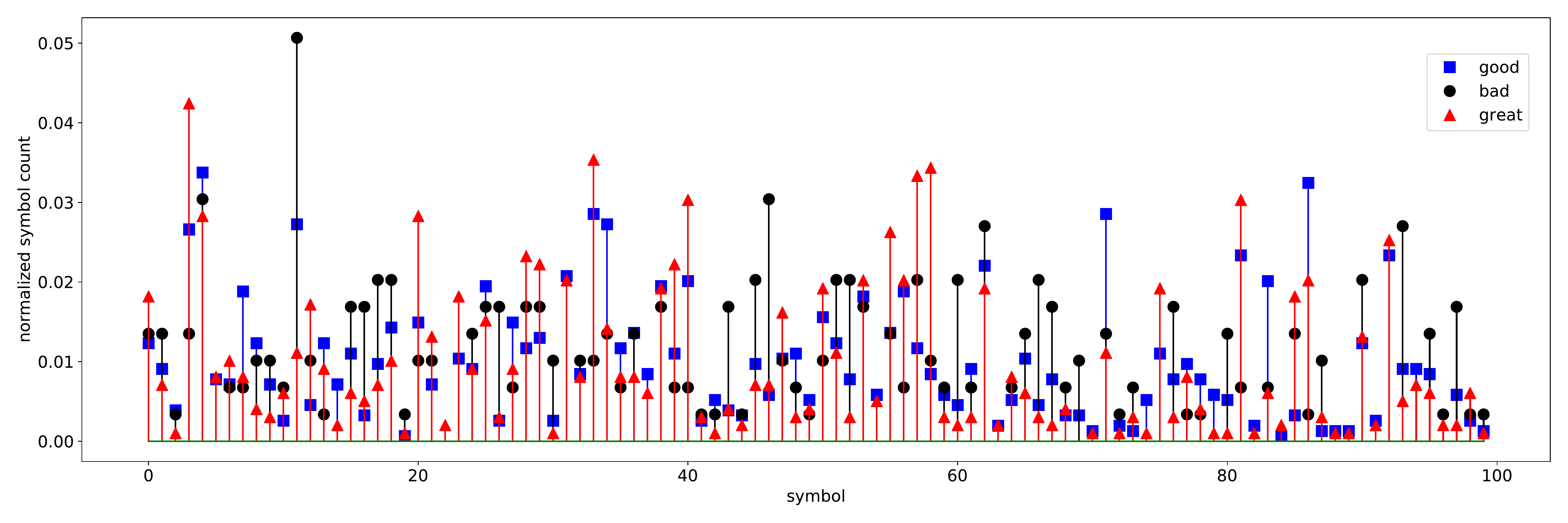}
\caption{Layer 12}
   \label{fig:attn_avg_layer12}
\end{subfigure}

\caption{Distribution of most attended symbols when generating tokens \textit{good}, \textit{bad}, \textit{great} for HSN(100, 5) trained on the Yelp data set. The decoder attention matrices between symbols and output are averaged over all attention heads for layers 1, 5 and 12.}
\label{fig:attn_avg}
\end{figure*}

\begin{figure*}
\centering
\includegraphics[width=.7\linewidth]{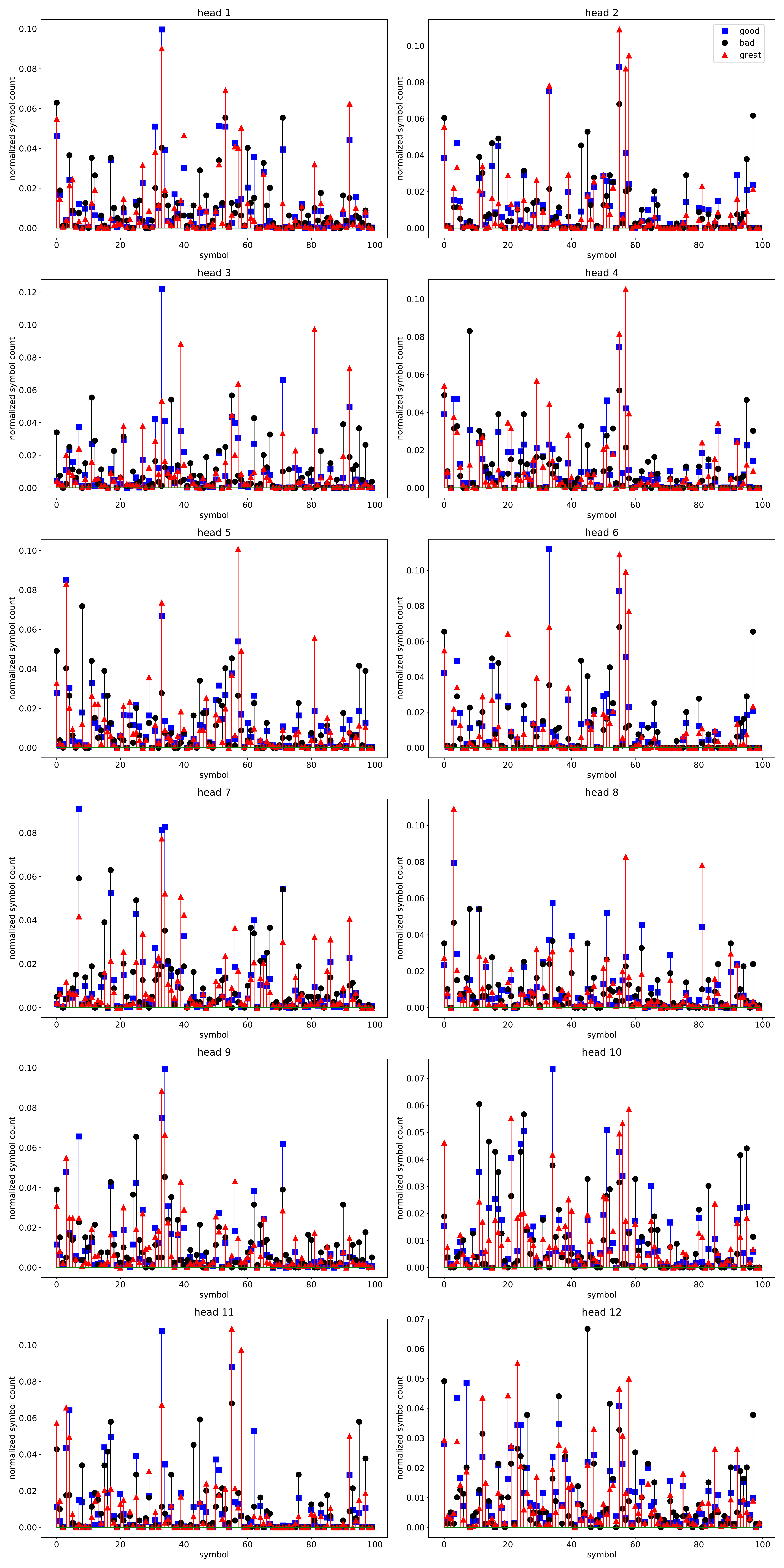}
\caption{Distribution of most attended symbols when generating tokens \textit{good}, \textit{bad}, \textit{great} for HSN(100, 5) trained on the Yelp data set. The distribution is computed from the decoder attention matrices between symbols and output for each attention head for layer 1.}
\label{fig:attn_by_head_layer1}
\end{figure*}

\begin{figure*}
\centering
\includegraphics[width=.7\linewidth]{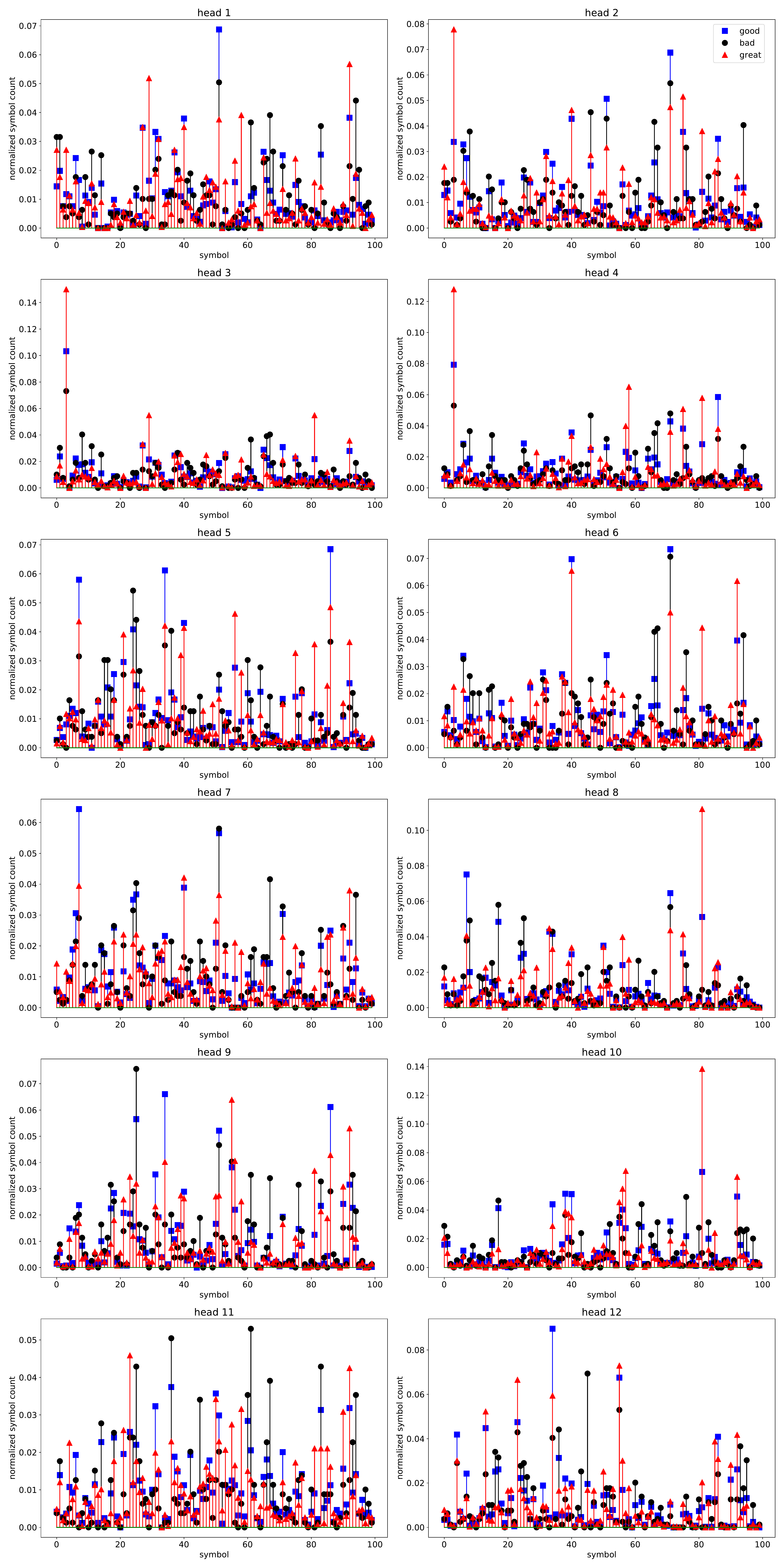}
\caption{Distribution of most attended symbols when generating tokens \textit{good}, \textit{bad}, \textit{great} for HSN(100, 5) trained on the Yelp data set. The distribution is computed from the decoder attention matrices between symbols and output for each attention head for layer 5.}
\label{fig:attn_by_head_layer5}
\end{figure*}

\begin{figure*}
\centering
\includegraphics[width=.7\linewidth]{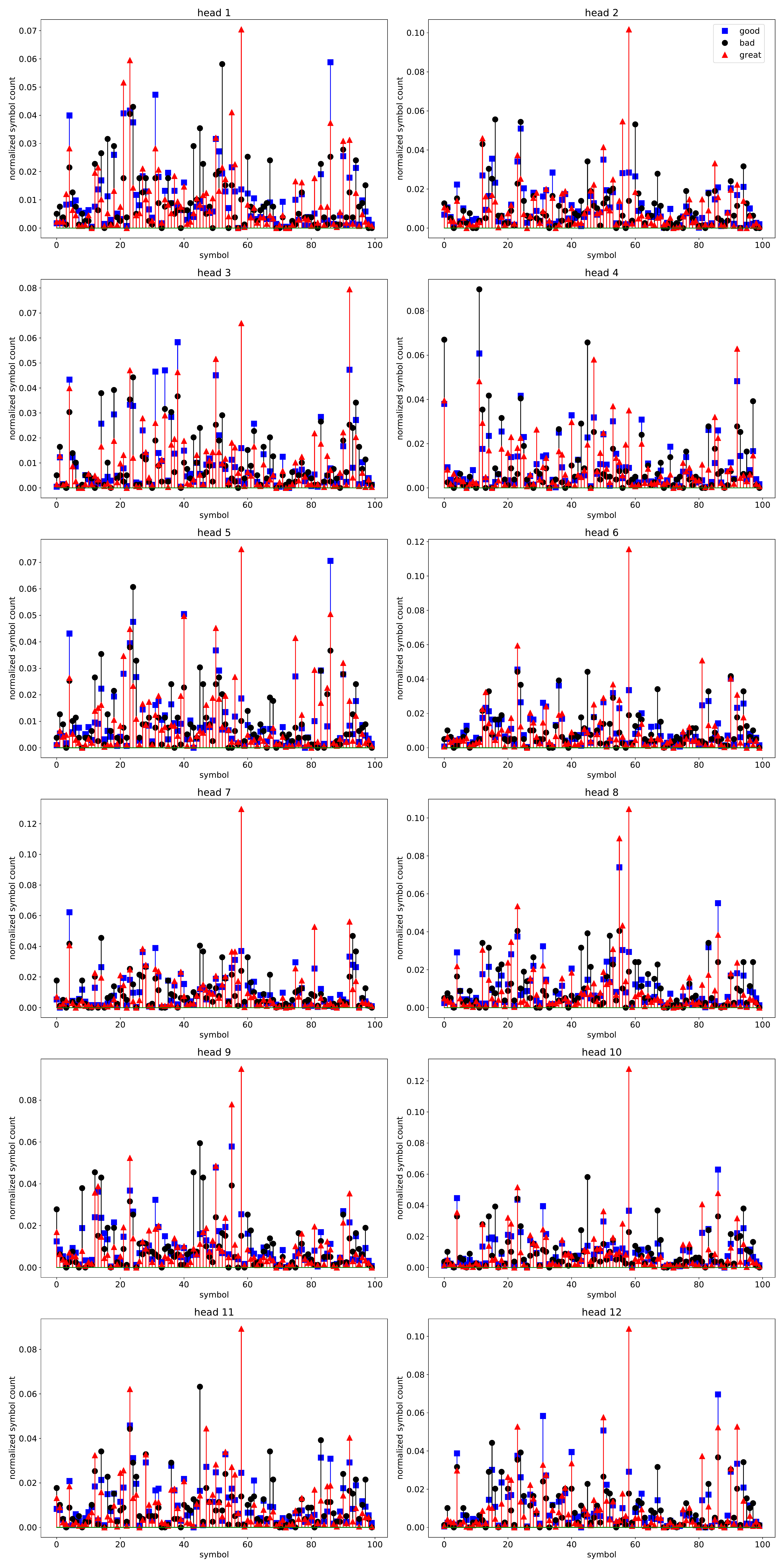}
\caption{Distribution of most attended symbols when generating tokens \textit{good}, \textit{bad}, \textit{great} for HSN(100, 5) trained on the Yelp data set. The distribution is computed from the decoder attention matrices between symbols and output for each attention head for layer 12.}
\label{fig:attn_by_head_layer12}
\end{figure*}

Thus, for a model trained on the Yelp dataset, we examine to which symbols does the decoder of HSN attend to, when processing the words \textit{good}, \textit{great} and \textit{bad}.
Figure \ref{fig:attn_avg} shows the most attended symbol distribution for layers 1 (first), 5 (middle), 12 (last), when averaging the attention matrices over all attention heads. Figures \ref{fig:attn_by_head_layer1}, \ref{fig:attn_by_head_layer5}, \ref{fig:attn_by_head_layer12} show these distributions for each head separately.
Note how, for a fixed token, the distribution of attention changes as one moves between heads and layers, although there are some repeating patterns too.

We can quantify these features by computing the Kullback-Leibler (KL) divergence between these distributions. 
The KL values are shown in Table \ref{tab:attn_kl}. 

Interestingly enough, the distribution of symbols that are attended to when processing the word \textit{great} is closer to the distributions of symbols attended by the word \textit{good}, than to the distributions of symbols attended by the word \textit{bad}.

\section{On Commonsense Reasoning Generation} \label{app:commonsense}

In this section we expatiate on the details of our approach to commonsense reasoning generation.  

 First, we modify the encoder component of HSN to process the tuples $(\*s, \*r, \*o)$ as
\begin{multline}
 \small
     q_{\phi}(\*z_{1:L}|[\*s, \*r, \*o], \*A) = \\
     q_{\phi}(\*z_{1:\frac{L}{2}}| [\*s, \*r], \*A) q_{\phi}(\*z_{\frac{L}{2}+1:L}| [\*s, \*r, \*o], \*z_{\frac{L}{2}}, \*A), \nonumber
 \end{multline}
so that the first half of the schema depends on subject and relation only, whereas the second half depends on the entire 3-tuple. 
As it will become evident below, this decoupling is necessary for the inference of novel objects.

Each of the posterior distributions above is modelled with the same architecture, as shown in Fig.~\ref{fig:model}, but sharing a single pretrained BERT model. That is, we have two copies of all pink-shaded blocks in the Fig. \ref{fig:model}, one for $q_{\phi}(\*z_{1:\frac{L}{2}}| [\*s, \*r], \*A)$, the other for $q_{\phi}(\*z_{\frac{L}{2}+1:L}| [\*s, \*r, \*o], \*z_{\frac{L}{2}}, \*A)$, and a single pretrained BERT model.
Using such a 2-component encoder model we are able to successfully infer schema representations for the KG tuples, as shown in Table \ref{tab:atomic}.
The task is however to infer \textit{new objects}, given only  subject-relation pairs. We thus need a way to infer schema representations without relying on the phrase object.

The classical solution to this inference problem is to replace, \`a la Kalman Filter, $q_{\phi}(\*z_{\frac{L}{2}+1:L}| [\*s, \*r, \*o], \*z_{\frac{L}{2}}, \*A)$ with a local, trainable prior model of the form $p_{\theta}(\*z_{\frac{L}{2}+1:L}| [\*s, \*r], \*z_{\frac{L}{2}}, \*A)$ -- where $\*z_{\frac{L}{2}}$ is sampled from $q_{\phi}(\*z_{1:\frac{L}{2}}| [\*s, \*r], \*A)$ --
and train the prior via the KL term in Eq.~\ref{eq:ELBO}. 
However, and as shown in Section \ref{app:training_objective}, maximizing the mutual information between data and representations averages out all local information in the KL term, and thus hinders the learning of the prior.

An alternative is to train, in the spirit of knowledge distillation \citep{44873}, a third-party model on the inferred schemata, to predict $\*z_{\frac{L}{2}+1:L}$ conditioned on $\*z_{1:\frac{L}{2}}$. 
Indeed, given the inferred schemata from the training KG, we consider a sequence-to-sequence model which inputs $\*z_{1:\frac{L}{2}}$, together with the subject-relation pair, and outputs $\*z_{\frac{L}{2}+1:L}$. 
That is, a model of the form $p_{\theta}(\*z_{\frac{L}{2}+1:L} | \*z_{1:\frac{L}{2}}, [\*s, \*r])$.
Specifically we use (i) a bidirectional LSTM network with hidden dimension of 512 to encode the first half of the schemata, (ii) a pretrained BERT model to encode the subject-relation pair, and (iii) a LSTM network of dimension 512 as an autoregressive decoder model. The initial (hidden) states of the latter are determined by an MLP which inputs the representations from the LSTM and BERT encoder models. The model is trained on samples from $q_{\phi}(\*z_{\frac{L}{2}+1:L}| [\*s, \*r, \*o], \*z_{\frac{L}{2}}, \*A)$.

Our preliminary results, $\text{HSN}_{ \tiny \text{(50, 20)}}^{\tiny \text{AR}}$ in Table \ref{tab:atomic}, are comparable to a stand-alone COMET(GPT-2) model, even when the thrid-party model only learns about $58\%$ of the second half of the schemata.
 \subsection{Atomic Dataset}

For this preliminary study we focus only on the ATOMIC dataset of \citet{sap2019atomic}.
It contains 877K $(\*s, \*r, \*o)$ tuples covering a variety
of social commonsense knowledge around specific
\textit{If-Then} events. 
A bit more in detail, ATOMIC splits its commonsense knowledge into nine
categories, covering the event’s causes, its effects on the agent, and its effect on other direct (or
implied) participants.
We use the training splits
from \citet{sap2019atomic}, resulting in 710K training,
80K validation, and 87K test tuples respectively.

\section{Training Time and Resource Consumption}\label{app:training_times}
In Table \ref{tab:train_times} we report the training times of our HSN models. The times are similar for all HSN configurations, so we report their average training time for each dataset.
\begin{table}[th]
    \centering
    \begin{tabular}{l c c c} 
    \hline
    dataset & train time & \#sentences & avg words \\
    & in hours & & per sentence \\
    \hline
    Yahoo & 22 & 100K & 80 \\
    Yelp & 26 & 100K & 97 \\
    PTB & 5 & 38K & 22 \\
    Atomic & 42 & 877 & 6 \\
    \hline
    \end{tabular}
    \caption{Average training times of all HSN configurations per dataset. All HSN models have about 252M parameters. We ran each model on a single NVIDIA A100-SXM4-40GB GPU unit.
}
    \label{tab:train_times}
\end{table} 
\begin{figure*}[th!]
\centering
\begin{subfigure}[b]{\textwidth}
   \includegraphics[width=1\linewidth]{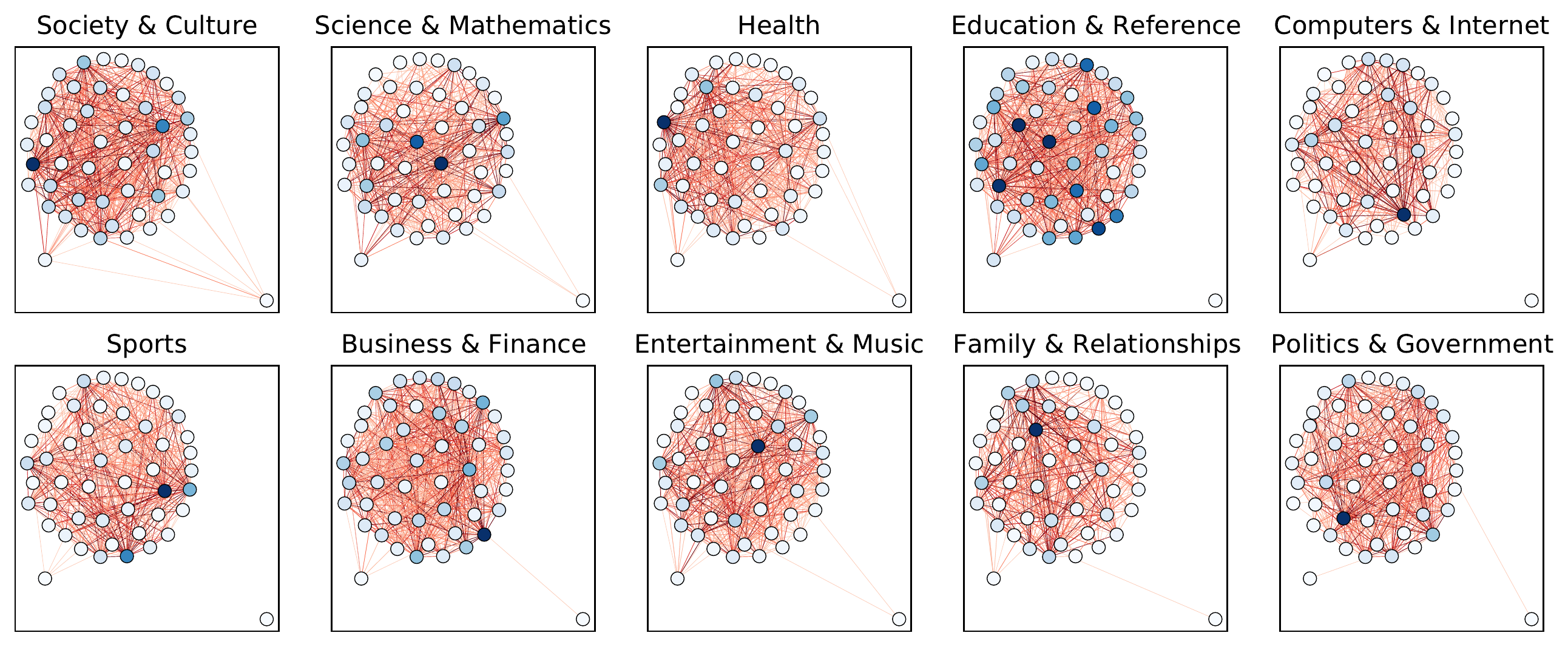}
  \caption{HSN(50, 5)}
   \label{fig:yahoo-HSN-50-5} 
\end{subfigure}

\begin{subfigure}[b]{\textwidth}
   \includegraphics[width=1\linewidth]{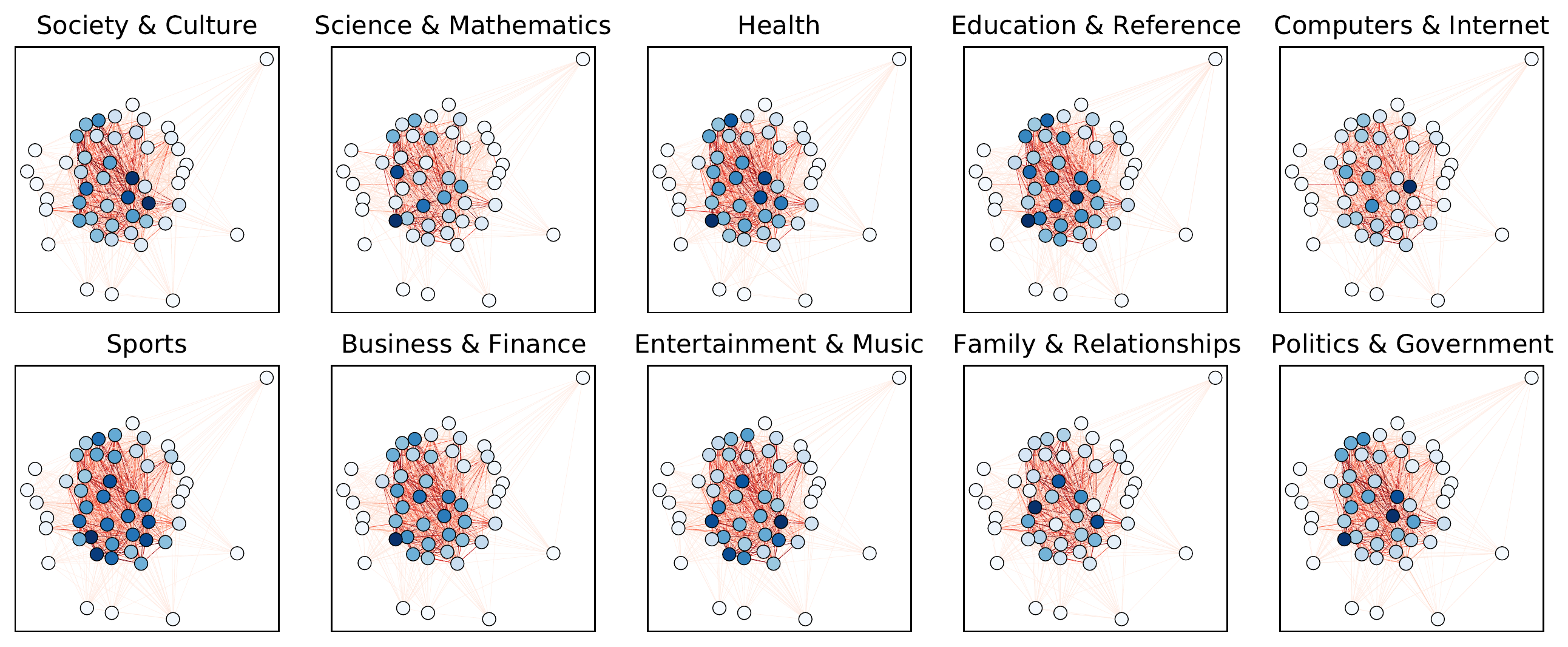}
\caption{HSN(50, 20)}
   \label{fig:yahoo-HSN-50-20}
\end{subfigure}
\caption{Schema distributions inferred from each category of the Yahoo dataset, for HSN(50, $L$) with $L=\{5, 20\}$. The node positions in the figure are consistent among labels and were computed using a force-directed embedding of the global graph ${\cal G}$.}
\label{fig:yahoo-50}
\end{figure*}

\begin{figure*}[th!]
\centering
\begin{subfigure}[b]{\textwidth}
   \includegraphics[width=1\linewidth]{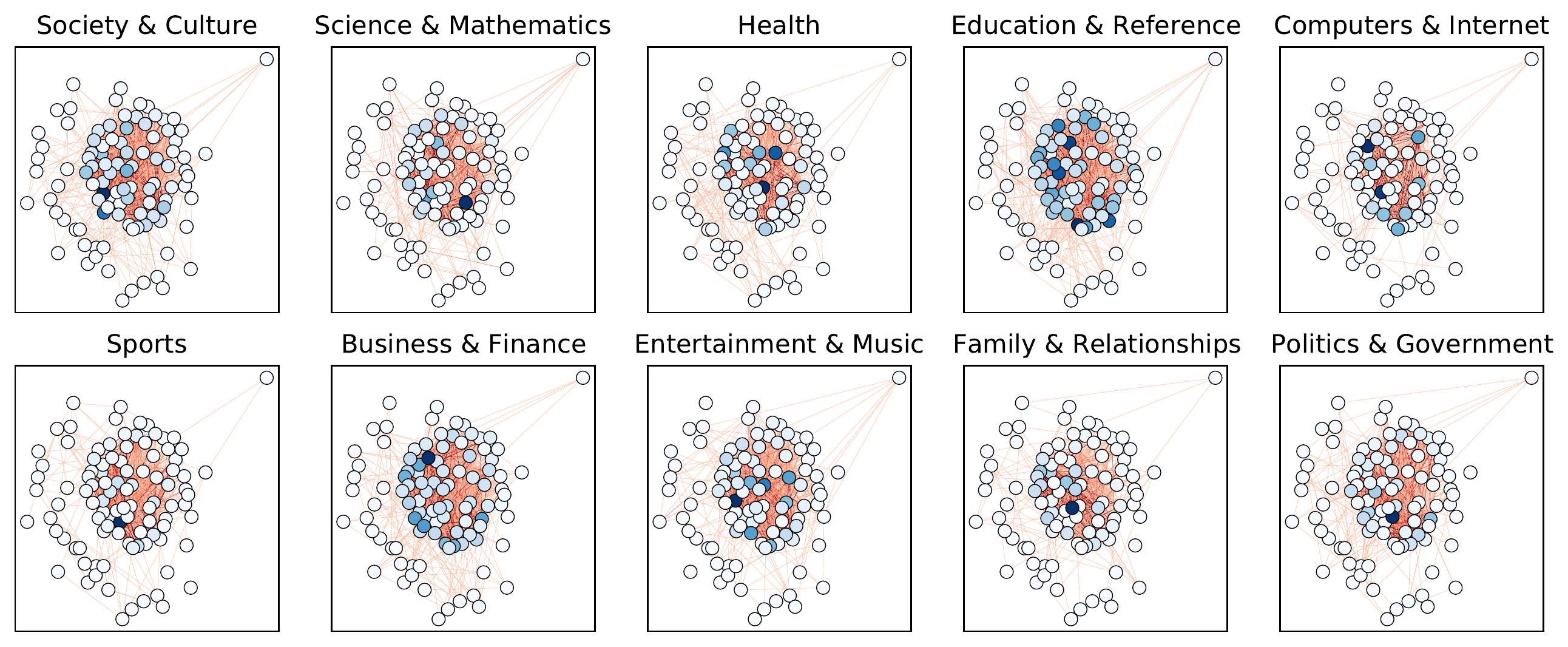}
  \caption{HSN(100, 5)}
   \label{fig:yahoo-HSN-100-5} 
\end{subfigure}

\begin{subfigure}[b]{\textwidth}
   \includegraphics[width=1\linewidth]{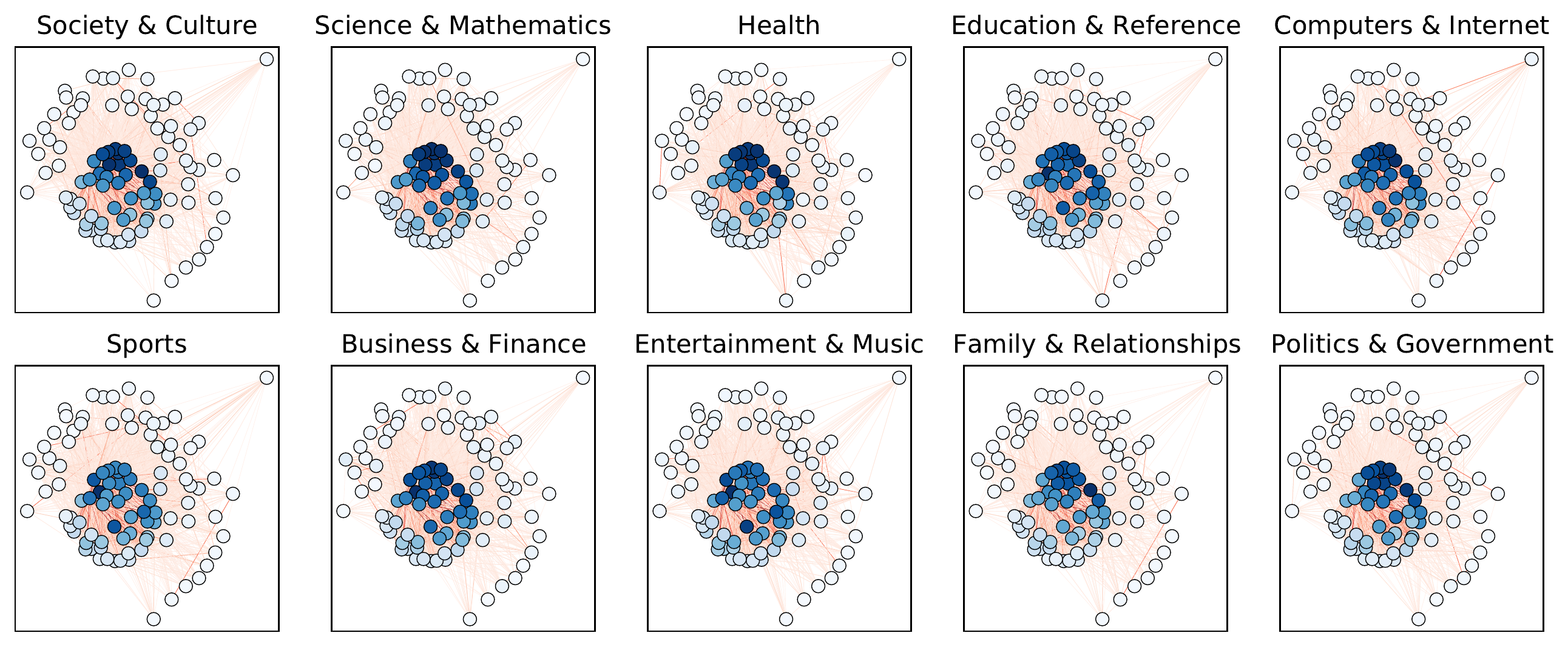}
\caption{HSN(100, 20)}
   \label{fig:yahoo-HSN-100-20}
\end{subfigure}
\caption{Schema distributions inferred from each category of the Yahoo dataset, for HSN(100, $L$) with $L=\{5, 20\}$. The node positions in the figure are consistent among labels and were computed using a force-directed embedding of the global graph ${\cal G}$.}
\label{fig:yahoo-100}
\end{figure*}

\begin{figure*}[th!]
\centering
\begin{subfigure}[b]{\textwidth}
   \includegraphics[width=1\linewidth]{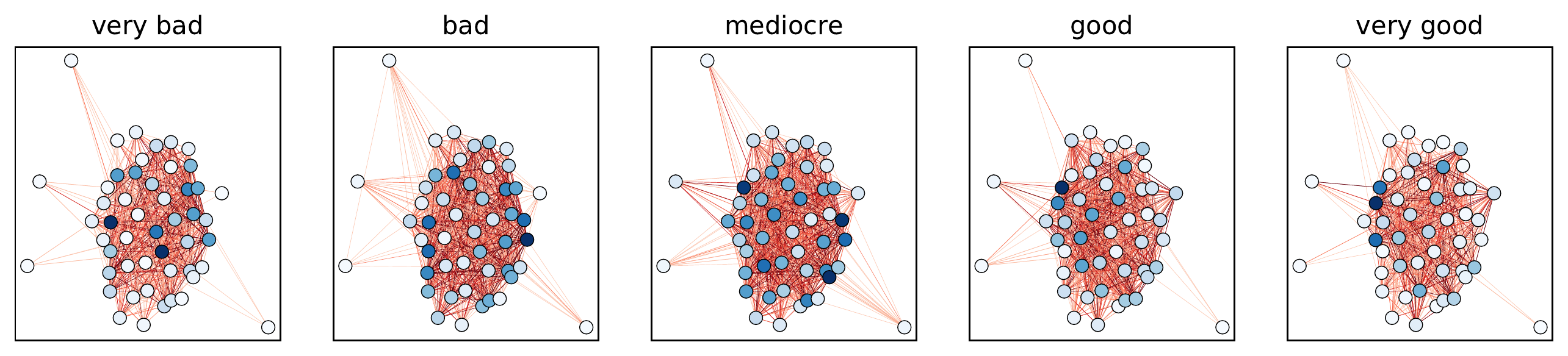}
  \caption{HSN(50, 5)}
   \label{fig:yelp-HSN-50-5} 
\end{subfigure}
\begin{subfigure}[b]{\textwidth}
   \includegraphics[width=1\linewidth]{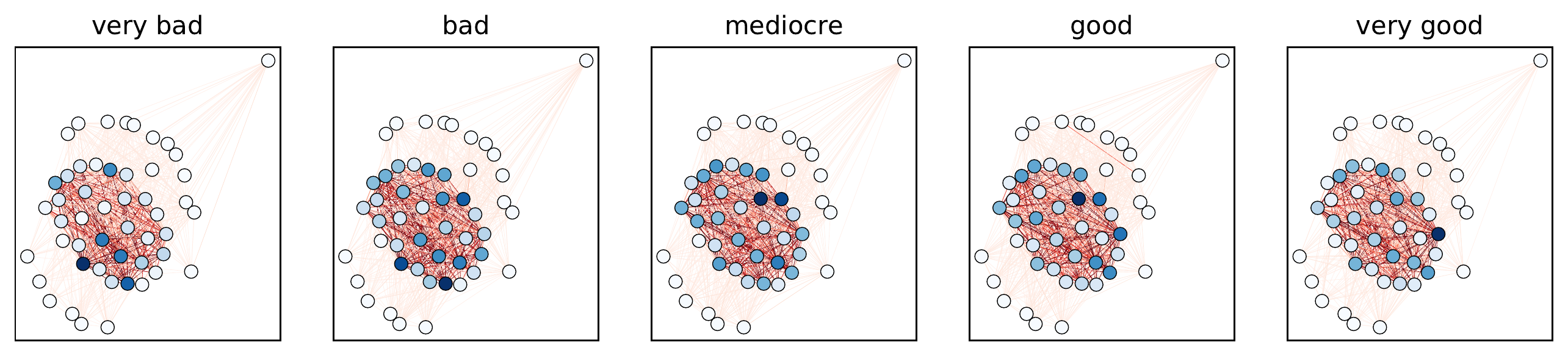}
\caption{HSN(50, 20)}
   \label{fig:yelp-HSN-50-20}
\end{subfigure}
\begin{subfigure}[b]{\textwidth}
   \includegraphics[width=1\linewidth]{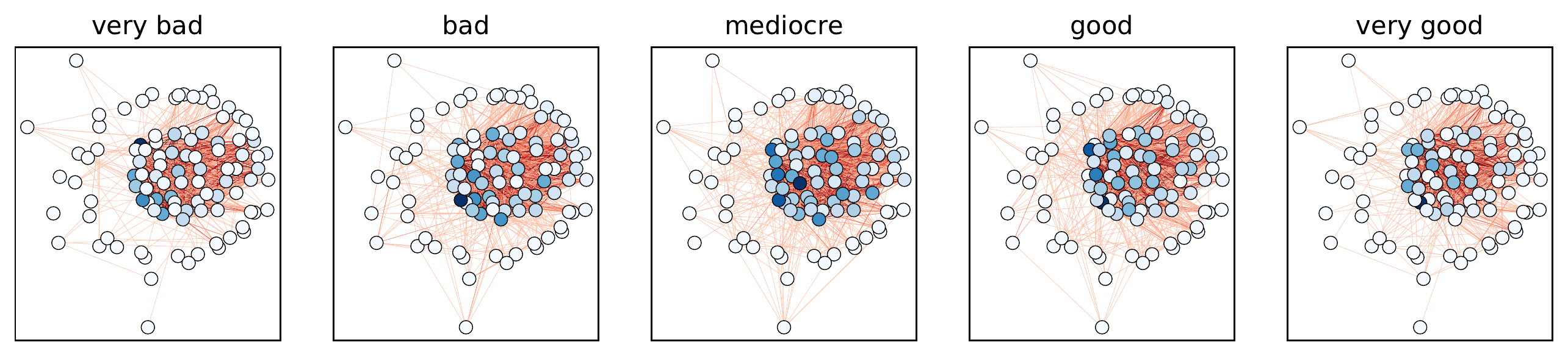}
\caption{HSN(100, 5)}
   \label{fig:yelp-HSN-100-5}
\end{subfigure}
\begin{subfigure}[b]{\textwidth}
   \includegraphics[width=1\linewidth]{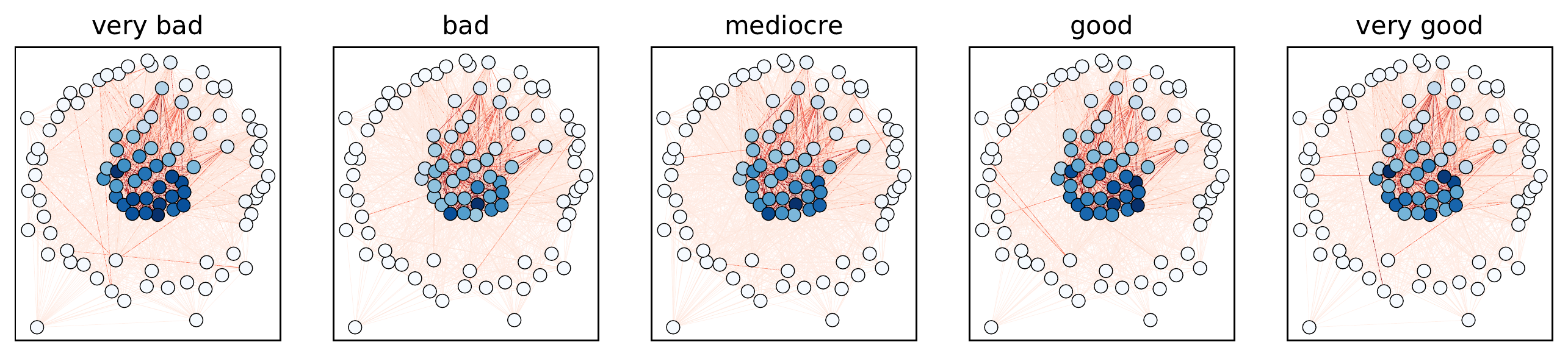}
\caption{HSN(100, 20)}
   \label{fig:yelp-HSN-100-20}
\end{subfigure}
\caption{Schema distributions inferred from each category of the Yelp dataset. The node positions in the figure are consistent among labels and were computed using a force-directed embedding of the global graph ${\cal G}$.}
\label{fig:yelp}
\end{figure*}


\newpage

\begin{table}[ht!]
\footnotesize
\begin{tabular}{m{.90\textwidth}}
Interpolate Society \& Culture Science \& Mathematics \\ \hline

\textbf{\noindent
is steady eye contact good? when i am communicating with someone, i tend to give 
very steady, long \_UNK eye contact.  
so i tried to \_UNK it as a young naive girl, and now  it's a habit i can't lose...
it just depends on the person you are having the conversation with... }

\begin{enumerate}[label=(\arabic*), noitemsep, topsep=0pt]
\item i am 14 years old. their is this girl again who speaks very much of me and talks 2 me, the idea of  me 2 her and never gives any suggestion to verbal \_UNK for my 2nd. listen, therapy!
\item what do you do when you think your best friend told you shes bisexual? when she says that,   or you might have believed if your friend said it is. they're inevitably sharing that they don't share... 
\item how do you change liquid in an ice cube into liquid form? paste, mix and freeze \_UNK a gallon of co2  into a \_UNK and then \_UNK in some ice to form a coating. 
\item what kind of rules does gravity apply? if a certain weight is placed in a container, the net force  applied on it hits the water surface and the right weight will turn into gravity 
\end{enumerate}
\textbf{how does a photovoltaic system that feeds back into the power grid get on the same phase angle?  or? should i say does it need to be the same as the \_UNK's? }\\
\hline
\\ \\
Interpolate: Business \& Finance Family -- Relationships \\ \hline
\textbf{\noindent
at 35, am i too old to go to college to become a psychiatrist? i'm 37, and i just started my  second semester in a 2 year college... you need to be prepare for the financial aspects, but  the social ones are no problem...
}
\begin{enumerate}[label=(\arabic*), noitemsep, topsep=0pt]
\item what would be a good title for a \_UNK \_UNK? i have \_UNK in \_UNK and there are no real courses done  for it but i do love the job and i've already done my freshman year. i am currently teaching placement  at \_UNK and need the same as the average undergraduate student... 
\item has anyone here applied in the past 4 months or is it better to get a try out y \_UNK a slightly better  long term career 
\item lately im having trouble with my fiancee, how do i bring him back? it obvious at this point that you  can't `` bring us together ''. try playing games. 
\item could i still go out with this guy and still be friends and respect him.? i don't want to just fell in love with the guy that i was with. i want 2 be with friend's girl and still be friends...
\end{enumerate}
\textbf{  how do i know if my man, is inlove with me? well... some questions, how old are \_UNK?  - are you wealthy?, is he wealthy?, how long have you been together... } \\
\hline
\end{tabular}
\caption{\footnotesize Interpolation between four random instances from the Yahoo dataset}
\label{tab:interpol-supmat-2}
\end{table} \begin{table}[ht]
\footnotesize
\begin{tabular}{m{0.90\textwidth}}

Interpolate: Very bad - bad\\ \hline

\textbf{\noindent 
do not use this company!! they told me within one hour, then i called again they said the driver have 90 mins. 90 minutes later, they said the driver is in traffic and wait for 15 minutes, i checked google map no accident, all green on all freeway...
} 
\begin{enumerate}[label=(\arabic*), noitemsep, topsep=0pt]
\item i ordered for pick up as my daughter hadn't been told that or even ordered online. when i  spoke to the young  lady, who was \_UNK, she carried on a conversation with not a manager. it's bad customer service and  i wouldn't even bother with this place...
\item place was clean... when i called to let them know i 'd get something else, the person that answering the phone wouldn't understand me... really? i gave this restaurant a b + for the cleanliness of the food  and the friendliness of the staff
\item i had the quesadilla and the carnitas tacos. i felt every bite of these were so rubbery and the potatoes were off. i feel like the service and the quality of food can do much better. 
\item somewhat disappointed. i did it once and loved it but today, today's water is bitter and salty... and the mint and cherry blossom \_UNK'flavors just taste that way. 
\end{enumerate} 
\textbf{the food quality doesn't match the place at all. i think it's ok for a pub but this place is supposed  to be a nice place for professional lunches. i had the chicken flatbread and the chicken was more  like subway chicken! with so many options around that area i won't pick this place for lunch. } \\
\hline
\\

\\
Interpolate: Very bad -- Very Good \\ \hline
\textbf{\noindent skip it... there are much better options out there! the `` hot '' food was not hot, and the flavor  was only mediocre at most.
} 
\begin{enumerate}[label=(\arabic*), noitemsep, topsep=0pt] \item indifferent to locals. the kids size pizzas were a billion times worse than a pizza hut. the quality of food   was just awful. i wouldn't recommend this to a significant other for what it is. \item this new mexican spot is ok, bordering on childish. i went with friends and ordered a carne asada burro... it wasn't off the hook ; what made this place great were the chips \& salsa sucked. yuck! ... \item wow. \_UNK you give so much frosting!! we were a groupon special for a cupcake for the princess of chocolate,  and we were pretty stoked. they were \_UNK and creative. they even suggested we try the coconut ... 
we 'll definitely be back soon.
\item went for the first time during a recent trip to vegas. our server jeff made special recommendations for our friends and i. it was fantastic most of the food was light and fresh...i would highly recommend this place!
\end{enumerate}
\textbf{i had dinner at republic kitchen tonight for the first time and was very impressed with the service,  the decor, the menu, and the food quality...
i am going back sunday for their brunch and jazz!  } \\
\hline
\end{tabular}
\caption{\footnotesize Interpolation between four random instances from the Yelp dataset}
\label{tab:interpol-supmat-3}
\end{table}

\end{document}